\documentclass{article}

\PassOptionsToPackage{numbers,sort&compress}{natbib}
\usepackage[preprint]{neurips_2026}
\makeatletter
\renewcommand{\@noticestring}{}
\renewcommand{\@notice}{}
\makeatother

\usepackage{graphicx}
\usepackage{wrapfig}

\usepackage[utf8]{inputenc}
\usepackage[T1]{fontenc}
\usepackage{hyperref}
\usepackage{url}
\usepackage{booktabs}
\usepackage{amsfonts}
\usepackage{nicefrac}
\usepackage{microtype}
\usepackage{xspace}
\usepackage{xcolor}

\usepackage{multirow}
\usepackage[table]{xcolor}
\usepackage{tabularx}
\usepackage{algorithm}
\usepackage{algpseudocode}
\usepackage{float}
\usepackage{placeins}
\usepackage{dblfloatfix}
\usepackage{amsmath}
\usepackage{makecell}
\usepackage{enumitem}

\usepackage{CJKutf8}

\usepackage{pifont}
\newcommand{\cmark}{\ding{51}}
\newcommand{\xmark}{\ding{55}}

\newcommand{\name}{\texttt{\textbf{DGP}}\xspace}

\definecolor{mathbg}{RGB}{252,242,242}
\definecolor{codebg}{RGB}{241,247,252}
\definecolor{improvebg}{RGB}{225,242,225}
\definecolor{sectionbg}{RGB}{242,242,242}

\title{Where to Connect? Boosting MLLMs via Dynamic Gated Pathways across ALL ViT and LLM Layers}

\author{
  \textbf{Yingying Yan}\textsuperscript{\normalfont 1,*} \quad
  \textbf{Jiaqi Tang}\textsuperscript{\normalfont 2,*} \quad
  \textbf{Wei Wei}\textsuperscript{\normalfont 1,\textdagger} \quad
  \textbf{Qianzhou Wang}\textsuperscript{\normalfont 1} \quad
  \textbf{Lei Zhang}\textsuperscript{\normalfont 1} \\[0.5em]
  \textbf{Jianmin Chen}\textsuperscript{\normalfont 1} \quad
  \textbf{Yuyang Xia}\textsuperscript{\normalfont 1} \quad
  \textbf{Botong Geng}\textsuperscript{\normalfont 1} \quad
  \textbf{Jinjian Wu}\textsuperscript{\normalfont 1} \quad
  \textbf{Qifeng Chen}\textsuperscript{\normalfont 2} \\[1em]
  \textnormal{\textsuperscript{1}Northwestern Polytechnical University, Xi'an, China} \\
  \textnormal{\textsuperscript{2}Hong Kong University of Science and Technology, Hong Kong, China} \\[0.75em]
  \textnormal{\texttt{\{yanyyy, wangqianzhou, chenjianmin, wujinjian\}@mail.nwpu.edu.cn}} \\
  \textnormal{\texttt{\{weiweinwpu, nwpuzhanglei\}@nwpu.edu.cn}} \\
  \textnormal{\texttt{jtang092@connect.ust.hk}, \texttt{cqf@ust.hk}}
}

\begin{document}

\maketitle
{
  \renewcommand{\thefootnote}{\fnsymbol{footnote}}
  \footnotetext[1]{Equal Contribution.}
  \footnotetext[2]{Corresponding author: Wei Wei.}
}

\begin{CJK*}{UTF8}{gbsn}

\begin{abstract}


A central yet under-explored question in Multimodal Large Language Models (MLLMs) is \emph{where to connect}---how visual information from a hierarchical vision encoder should be wired into the layer-wise semantics of a large language model. Mainstream MLLMs inject a fixed single-layer visual representation into the LLM and treat all LLM layers as a single consumer, ignoring that different layers demand different visual granularities. Recent multi-layer fusion methods either compress hierarchical features on the encoder side or rely on predefined sparse or hierarchical layer-to-layer connections, and therefore stop short of modeling \emph{adaptive} cross-layer interactions between the ViT and LLM hierarchies. We address this gap with \name (Dynamic Gated Pathways), a routing mechanism that establishes \emph{all-to-all} connections between every ViT layer and every LLM layer, with each LLM layer dynamically gating its preference over visual layers conditioned on its own semantic state. Built on top of LLaVA-1.5-7B and 13B, \name delivers consistent gains over the LLaVA-1.5-7B baseline (e.g., +4.2 on SQA$^{I}$, +3.3 on MM-Vet, +3.1 on VizWiz and LLaVA$^{W}$) and surpasses prior multi-layer visual fusion methods on most benchmarks. Layer-wise routing analyses further show that different LLM layers do select different visual granularities on demand, supporting both the effectiveness and interpretability of the proposed dynamic gated pathways. The code is available: \url{https://github.com/YingyingYan1/DGP}.

\end{abstract}

\section{Introduction}


Multimodal Large Language Models (MLLMs)~\cite{li2023blip,bai2025qwen3,li2025mini,hurst2024gpt} encode an image with a pretrained vision backbone~\cite{radford2021learning, oquab2023dinov2, zhai2023sigmoid}, project the resulting features through an adapter, and inject the visual tokens into a language model. In this paradigm, the effectiveness of visual information depends not only on the representational power of the encoder~\cite{yan2026heattok,chen2026remember,tang2026robust}, but---more critically---on \emph{where to connect}: how the visual tokens are wired into the internal layers of the LLM.

\begin{figure*}[t]
  \centering
  \includegraphics[width=1\linewidth]{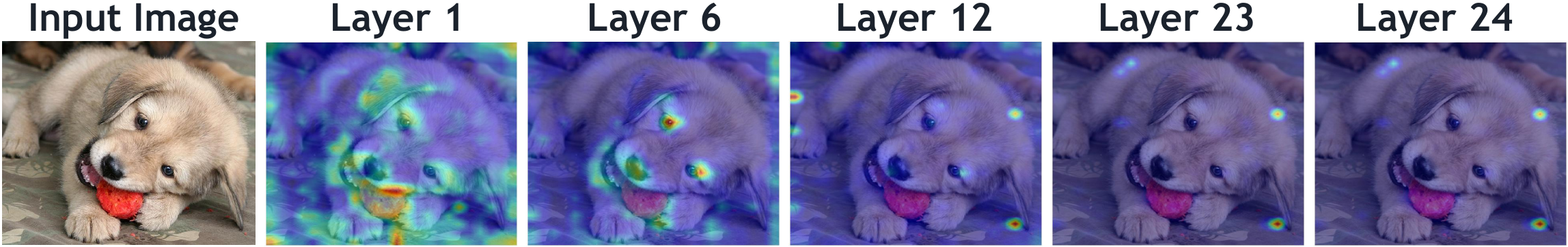}
  \vspace{-0.15in}
  \caption{\textbf{Attention distributions across ViT layers.} Shallow layers show dispersed attention over edges and textures; deeper layers progressively concentrate on a few semantic anchors.}
  \label{fig:attention_heatmap}
  \vspace{-0.06in}
\end{figure*}
\begin{figure*}[t]
  \centering
  \includegraphics[width=1\linewidth]{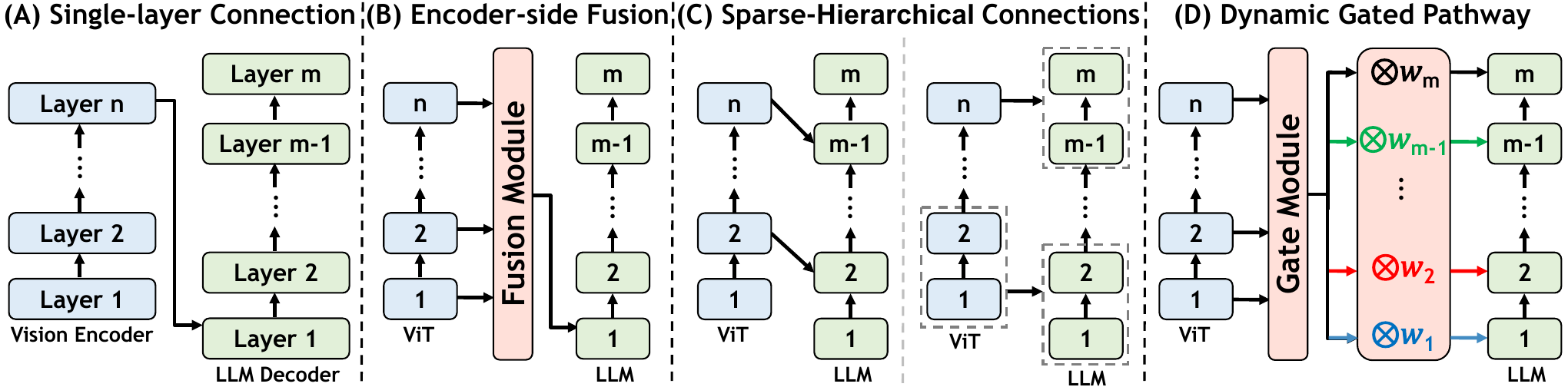}
  \vspace{-0.15in}
  \caption{\textbf{Strategies for visual token insertion.} (A) Fixed single-layer connection. (B) Encoder-side fusion of multiple ViT layers into one representation. (C) Predefined sparse or hierarchical connections between ViT and LLM layers. (D) \name (Ours): dynamic gated pathways between \emph{all} ViT and \emph{all} LLM layers. $w_m$ denotes the routing weights that LLM layer~$m$ assigns across \emph{all} ViT layers.}
  \label{fig:motivation}
  \vspace{-0.06in}
\end{figure*}

As illustrated in Figure~\ref{fig:attention_heatmap}, attention patterns across ViT layers differ markedly: shallow layers attend to edges, textures, and local details, whereas deeper layers progressively concentrate on a small number of semantic anchors. Yet, most MLLMs~\cite{bai2023qwen,liu2024improved,chen2024internvl} simply forward the output of one fixed ViT layer (typically the last or penultimate layer) to the LLM, as illustrated in Figure~\ref{fig:motivation}(A), effectively collapsing the image into a single-granularity semantic representation. Connecting the LLM only to the terminal deep layer therefore discards shallow- and mid-level cues and prevents the LLM's layer-wise semantics from interacting with the full visual hierarchy.

Two lines of work have started to expose multi-layer visual features to the LLM, but each falls short on the \emph{where to connect} question. \textbf{Encoder-side fusion}~\cite{yao2024dense, cao2024mmfuser, li2026instruction, lin2026text, yifanccra} merges multi-layer features into one representation before the LLM (Figure~\ref{fig:motivation}(B)); any dynamism is consumed inside the encoder and cannot adapt to the differentiated demands of individual LLM layers. \textbf{Predefined sparse or hierarchical connections}~\cite{meng2024deepstack, zhang2026sparse, wei2025dynamic, chen2024evlm} wire selected ViT layers to specific LLM layers (Figure~\ref{fig:motivation}(C)), typically using hand-crafted connection patterns or fixed hierarchical mappings. CLI~\cite{chen2026one} introduces adaptive gating conditioned on the current LLM state, but operates over sampled ViT layers and designated LLM injection points. Neither category explicitly models the \emph{adaptive}, \emph{dynamic} cross-layer interactions between \emph{all} ViT layers and \emph{all} LLM layers that the question of \emph{where to connect} demands.



Predictive coding in visual neuroscience suggests that perception is not a unidirectional feed-forward process but a hierarchical feedback loop that continuously modulates activations across layers~\cite{rao1999predictive}. We take this as inspiration to revisit visual token injection: visual evidence should not be delivered in one shot through a fixed pathway, but should keep \emph{participating} in the LLM's representation updates through dynamic cross-layer pathways. Building on this view, we introduce \name (Dynamic Gated Pathways), which connects \emph{all} ViT layers to \emph{all} LLM layers via learned, dynamically gated routes (Figure~\ref{fig:motivation}(D)).

Concretely, the outputs of different ViT layers act as candidate representations at varying granularities. At every LLM layer, \name produces a routing distribution over all ViT layers from the current hidden state, aggregates the candidates via gated weighted fusion, and injects the result back into the layer's visual hidden states through a residual connection. The hierarchical visual evidence keeps supplying cues for language reasoning, while the LLM's layer-wise semantic state drives the preference over visual hierarchies, yielding a cross-layer, progressively evolving vision--language interaction that integrates low-level details and high-level semantics \emph{on demand} at different decoding depths.


We instantiate \name on top of LLaVA-1.5~\cite{liu2024improved} and further validate it on Qwen3-VL-2B~\cite{bai2025qwen3} and LLaVA-NeXT~\cite{liu2024llavanext}. Across a wide range of multimodal benchmarks, \name yields consistent improvements over baselines and outperforms prior multi-layer visual feature utilization methods. The learned routing patterns confirm that different LLM layers indeed prefer different visual granularities, alleviating the information bottleneck of fixed single-layer fusion. Our contributions are:

\begin{itemize}[itemsep=3pt, topsep=0pt, parsep=0pt]
    \item We reframe visual information injection in MLLMs around the question of \emph{where to connect}, and show that fixed single-layer inputs and predefined sparse or hierarchical connections cannot match the differentiated demands of hierarchical ViT representations and layer-wise LLM semantics.
    \item We propose \name (Dynamic Gated Pathways), an all-to-all cross-layer connection mechanism in which every LLM layer dynamically gates its preference over every ViT layer conditioned on its current semantic state.
    \item We demonstrate the effectiveness and interpretability of \name on multiple multimodal benchmarks, where layer-wise routing analyses reveal the adaptive, layer-dependent selection behavior of the proposed dynamic gated pathways.
\end{itemize}

\section{Related Work}

\subsection{Multimodal Large Language Model}


A typical Multimodal Large Language Model (MLLM) couples a vision encoder, a vision--language adapter, and a large language model: the encoder turns the image into visual tokens, the adapter projects them into the language space, and the LLM consumes the joint visual--textual sequence to generate responses. Existing MLLMs mainly fall into two architectural families: projector-based decoder-only designs (e.g., LLaVA~\cite{liu2024improved}, MiniGPT-4~\cite{zhu2023minigpt}, Qwen-VL~\cite{bai2023qwen}, InternVL~\cite{chen2024internvl}) and cross-attention-based designs (e.g., Flamingo~\cite{alayrac2022flamingo}).


Current MLLMs use pretrained vision backbones such as CLIP-ViT~\cite{radford2021learning} and SigLIP~\cite{zhai2023sigmoid}, all of which expose representations of varying granularities across their layers. Yet most pipelines project only a single fixed layer into the LLM, so the language model sees only one level of visual abstraction. The visual connection mechanism is not merely a cross-modal interface: it determines whether the rich, hierarchical visual representations can be \emph{accessed and reused} inside the LLM.

\subsection{Connection Strategies for Visual Token Insertion}


Different ViT layers carry different semantic granularities: shallow layers retain local textures and spatial details, while middle and deep layers progressively encode object structures and high-level semantics~\cite{chen2025multimodal}. A single deep layer is not always the best choice~\cite{chen2025multimodal}, and exploiting the visual hierarchy depends as much on \emph{how} layers are connected to the LLM as on \emph{which} layers are selected~\cite{lin2025multi}. The crux, in other words, is to establish effective connections between vision-encoder layers and LLM layers---i.e., the \emph{where to connect} problem. Prior efforts can be grouped into two categories.

\textbf{Unified Encoder-side Connections.}
Methods in this category fuse multi-layer visual features on the encoder side and pass a single unified representation to the LLM. DenseConnector~\cite{yao2024dense} concatenates features from different ViT layers along the token or channel dimension. MMFuser~\cite{cao2024mmfuser} uses deep-layer features as queries to pull fine-grained information from shallow layers. IGVA~\cite{li2026instruction} and TGIF~\cite{lin2026text} adjust ViT-layer weights conditioned on the instruction. CCRA~\cite{yifanccra} combines several attention mechanisms for text-conditioned region--semantic alignment. While these designs improve multi-layer feature utilization, the entire fusion happens \emph{before} the LLM; the LLM still receives one homogenized input and cannot reach back into the visual hierarchy as decoding progresses.

\textbf{Predefined Sparse or Hierarchical Connections.}
Methods in this category route hierarchical visual information to different LLM layers. DeepStack~\cite{meng2024deepstack} distributes visual tokens across decoder depths according to a fixed hierarchical schedule. Sparse Shortcuts~\cite{zhang2026sparse} wires ViT and LLM layers through predefined connection patterns. EVLM~\cite{chen2024evlm} feeds multi-layer ViT features into different gated cross-attention layers. CLI~\cite{chen2026one} applies adaptive gating at sampled vision layers and designated LLM injection points, rather than treating every ViT--LLM pair as a candidate connection. A more recent line of work~\cite{wei2025dynamic} selects and fuses hierarchical visual features conditioned on each LLM layer's representation. Although these methods begin to acknowledge the hierarchical correspondence between vision and language, most still rely on hand-crafted connection patterns or fixed layer-to-layer mappings, and few of them simultaneously realize (i) \emph{all-to-all} cross-layer connectivity, (ii) routing weights driven by the \emph{current} LLM-layer state, and (iii) end-to-end training with a stable routing distribution---the design point that \name targets.


In summary, existing methods either compress the visual hierarchy into a single pre-LLM representation, or wire it into the LLM through predefined sparse or hierarchical connections; neither answers \emph{where to connect} via dynamic cross-layer interactions between \emph{all} ViT layers and \emph{all} LLM layers. We close this gap with \name, which establishes adaptive, all-to-all connections between ViT and LLM layers through dynamic gated pathways.

\section{Methodology}

\begin{figure*}[t]
  \centering
  \includegraphics[width=1\linewidth]{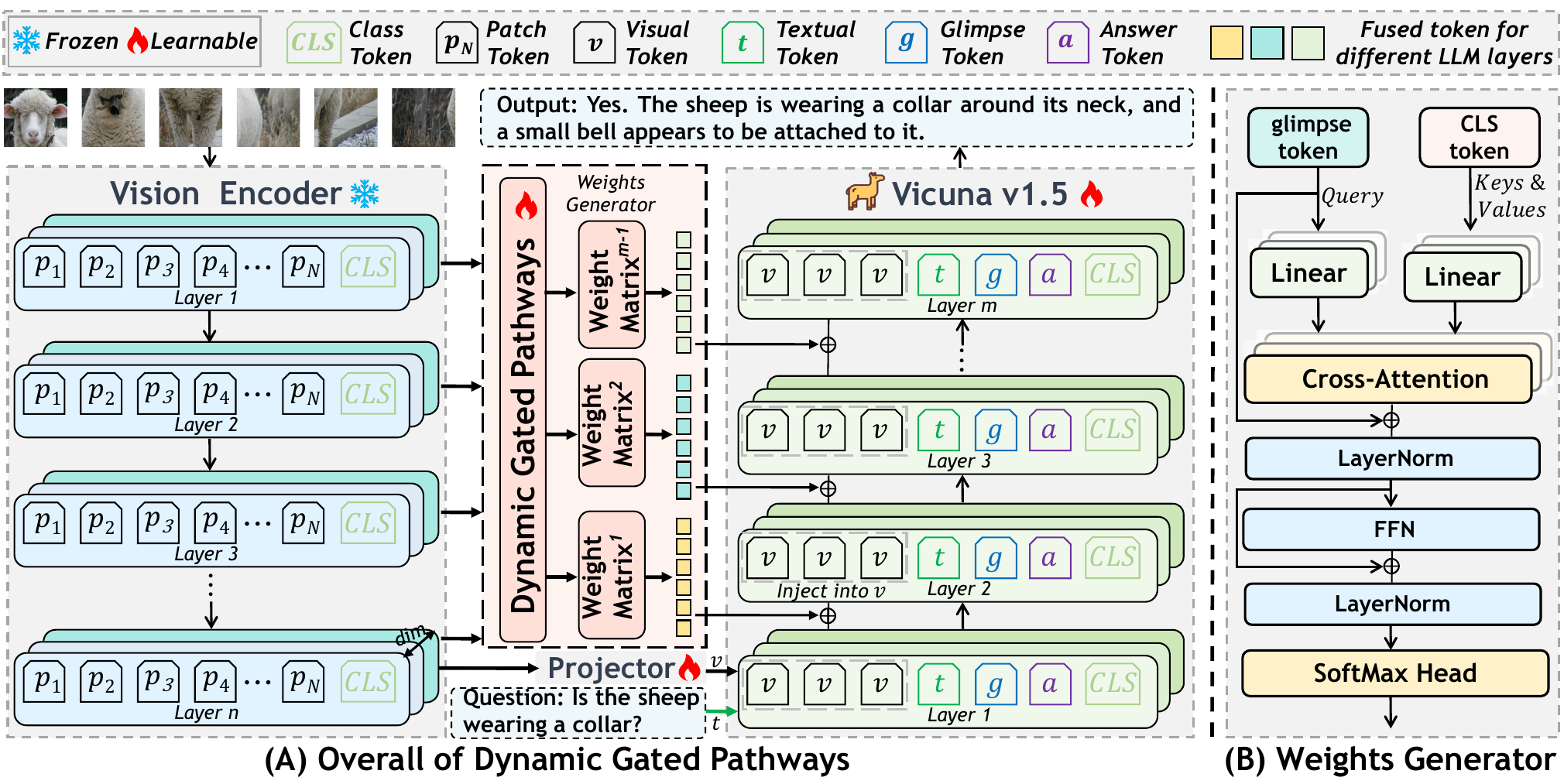}
  \vspace{-0.15in}
  \caption{\textbf{Overview of \name.} (A) At every LLM layer, the Dynamic Gated Pathways module routes over \emph{all} ViT layers conditioned on the current language-layer state, aggregates the selected visual features, and injects them back into the LLM hidden states through a gated residual connection---establishing all-to-all cross-layer connections between ViT and LLM. (B) Inside the module, ViT layer summaries serve as keys/values and the language-layer state as the query; a shared cross-attention followed by a layer-specific routing MLP produces the per-layer routing weights. \text{Weight Matrix}$^{(m)}$ denotes the routing weights of LLM layer $m$ over all ViT layers.}
  \label{fig:framework}
  \vspace{-0.06in}
\end{figure*}

\textbf{Problem Formulation.}
In existing MLLMs, a single fixed visual representation is injected once and then propagates implicitly through the LLM~\cite{bai2023qwen,liu2024improved,chen2024internvl}, leaving no direct cross-layer pathway between the visual hierarchy and the language hierarchy. As a result, different decoding stages cannot reach back to the visual granularity they actually need. \name (Dynamic Gated Pathways) removes this bottleneck by opening dynamic gated pathways between every ViT layer and every LLM layer, so that each LLM layer can adaptively pick the visual hierarchy that best fits its current semantic state.


Let the vision encoder have $L_v$ layers, with the output of the $\ell$-th layer ($\ell=1,\dots,L_v$) denoted as $\mathbf{X}_v^{(\ell)}$. The LLM has $L$ layers in total, and the hidden state of the $l$-th layer ($l=1,\dots,L$) is $\mathbf{H}^{(l)}\in\mathbb{R}^{(N_v+N_t)\times d}$, where $\mathbf{H}_v^{(l)}\in\mathbb{R}^{N_v\times d}$ corresponds to the visual token segment. For each LLM layer $l$, we aim to learn a layer-specific fusion function:
\begin{equation}
\widehat{\mathbf{Z}}^{(l)} = g^{(l)}\!\left(\mathbf{H}^{(l)}, \{\mathbf{X}_v^{(\ell)}\}_{\ell=1}^{L_v}\right),
\label{eq:gated_connection_function}
\end{equation}
where $\widehat{\mathbf{Z}}^{(l)}\in\mathbb{R}^{N_v\times d}$ is the fused visual representation tailored to the $l$-th LLM layer. The objective of $g^{(l)}(\cdot)$ is to allocate fusion weights over visual layers conditioned on the current language-layer state, so that visual evidence at multiple depths is integrated \emph{on demand} during decoding.

\subsection{All-to-All Cross-Layer Connections}

Our structural departure from typical MLLMs is to wire \emph{every} LLM layer ($l=1,\ldots,L$) to \emph{every} ViT layer ($\ell=1,\ldots,L_v$): each decoding depth retains direct access to the full visual hierarchy, rather than being restricted to one fixed depth or a hand-crafted sparse routing pattern.
As illustrated in Figure~\ref{fig:motivation}(D) and Figure~\ref{fig:framework}(A), we replace the conventional fixed single-layer visual injection with dynamic gated cross-layer pathways between the ViT and LLM stacks.
Given an input image $\mathbf{I}$, the vision encoder $F(\cdot)$ produces a hierarchy of visual representations:
\begin{equation}
\{\mathbf{X}_v^{(\ell)}\}_{\ell=1}^{L_v}=F(\mathbf{I}),
\qquad
\mathbf{X}_v^{(\ell)}
=
[\mathbf{c}^{(\ell)};\mathbf{P}^{(\ell)}],
\end{equation}
where $\mathbf{c}^{(\ell)}$ is the \texttt{[CLS]} representation of the $\ell$-th ViT layer and $\mathbf{P}^{(\ell)}\in\mathbb{R}^{N_v\times d_v}$ is its patch representation. Instead of relying on implicit propagation of the initial visual tokens, we connect every LLM layer to every ViT layer explicitly. Concretely, for any LLM layer $l=1,\ldots,L$, the model derives a layer-specific visual representation $\widehat{\mathbf{Z}}^{(l)}$ from $\{\mathbf{X}_v^{(\ell)}\}_{\ell=1}^{L_v}$ via Eq.~\eqref{eq:gated_connection_function}. This realizes an all-to-all connection set between visual and language hierarchies:
\begin{equation}
\mathcal{E}
=
\{(\ell,l)\mid \ell=1,\ldots,L_v,\; l=1,\ldots,L\}.
\end{equation}
Each LLM layer can therefore reach low-level details, mid-level structures, and high-level semantics directly, rather than being locked to a single visual layer or a hand-crafted sparse routing pattern.


The layer-specific representation $\widehat{\mathbf{Z}}^{(l)}$ is written back into the visual token segment of the same LLM layer via a gated residual connection (formalized in the Token Fusion paragraph below). Importantly, $g^{(l)}(\cdot)$ is \emph{not} a uniform average over visual layers: it is realized by the Dynamic Gated Pathways module described next, which adaptively controls the strength of each ViT--LLM connection.

\subsection{Dynamic Gated Pathways Module}

\textbf{Weight Generation.}
Inspired by existing glimpse-token designs~\cite{zeng2025glimpse}, we prepend a learnable glimpse token $\mathbf{g}\in\mathbb{R}^{d}$ immediately before the first answer token of every multimodal training sample. Because of the causal autoregressive structure, this token can attend to all preceding visual and textual context, and therefore acts as a compact query signal that summarises the current decoding depth. Let $\mathbf{h}_g^{(l)}\in\mathbb{R}^{d}$ denote the hidden state at the glimpse position in the $l$-th LLM layer; we use it as the routing query, $\mathbf{Q}^{(l)}=\mathbf{h}_g^{(l)}$.

For each ViT layer $\ell$, we take its global \texttt{[CLS]} representation $\mathbf{c}^{(\ell)}$ as the layer summary: since \texttt{[CLS]} already aggregates the whole image inside that layer, it provides a compact layer-level descriptor of the corresponding visual granularity~\cite{dosovitskiy2020image}. We map all $L_v$ summaries into the LLM hidden space and stack them into shared keys and values, $\mathbf{K}=\mathbf{V}=\mathrm{Stack}\!\left(\widetilde{\mathbf{c}}^{(1)},\ldots,\widetilde{\mathbf{c}}^{(L_v)}\right)\in\mathbb{R}^{L_v\times d}$.
A multi-head cross-attention~\cite{vaswani2017attention} module---shared across all LLM layers---then models the interaction between the current language-layer state and the visual-layer summaries, yielding a routing feature:
\begin{equation}
\mathbf{r}^{(l)}
=
\mathrm{MHA}_{\mathrm{shared}}
\!\left(
\mathbf{Q}^{(l)},\mathbf{K},\mathbf{V}
\right)
+
\mathbf{Q}^{(l)},
\end{equation}
where $\mathbf{Q}^{(l)}$ encodes the multimodal context of the current LLM layer and $\mathbf{K},\mathbf{V}$ summarize all candidate visual layers; together they determine how the $l$-th LLM layer should allocate connection strength across the ViT hierarchy.

Conditioned on this routing feature, the $l$-th LLM layer produces a routing distribution over all ViT layers through a layer-specific---i.e., \emph{not} shared---routing MLP:
\begin{equation}
\boldsymbol{\omega}^{(l)}
=
\mathrm{Softmax}
\!\left(
\mathbf{W}_2^{(l)}
\,
\mathrm{Dropout}
\!\left(
\mathrm{GELU}
\!\left(
\mathbf{W}_1^{(l)}\,\mathrm{LN}(\mathbf{r}^{(l)})+\mathbf{b}_1^{(l)}
\right)
\right)
+\mathbf{b}_2^{(l)}
\right)
\in\mathbb{R}^{L_v},
\end{equation}
where $\boldsymbol{\omega}^{(l)}=(\omega_1^{(l)},\ldots,\omega_{L_v}^{(l)})^{\top}$ denotes the routing weights of LLM layer $l$ over the $L_v$ ViT layers and satisfies $\sum_{k=1}^{L_v}\omega_k^{(l)}=1$. $\mathbf{W}_1^{(l)}\in\mathbb{R}^{h\times d}$ and $\mathbf{b}_1^{(l)}\in\mathbb{R}^{h}$ map the routing feature from the LLM hidden space to the routing hidden space; $\mathbf{W}_2^{(l)}\in\mathbb{R}^{L_v\times h}$ and $\mathbf{b}_2^{(l)}\in\mathbb{R}^{L_v}$ project it to an $L_v$-dimensional logit that softmax converts into the per-layer routing distribution. In short, $\mathrm{MHA}_{\mathrm{shared}}$ exposes the same query--summary interface to every LLM layer for fair comparison across visual layers, whereas the layer-specific routing MLP $\{\mathbf{W}_1^{(l)},\mathbf{b}_1^{(l)},\mathbf{W}_2^{(l)},\mathbf{b}_2^{(l)}\}$ lets each LLM layer voice its own preference over the visual hierarchy.

\textbf{Token Fusion.}
Given the routing weights, we fuse patch features from different visual layers position by position:
\begin{equation}
\widetilde{\mathbf{p}}_i^{(l)} = \sum_{k=1}^{L_v} \omega_k^{(l)} \mathbf{p}_i^{(k)}, \quad i=1,\dots,N_v,
\end{equation}
where $\mathbf{p}_i^{(k)}\in\mathbb{R}^{d_v}$ is the patch at position $i$ from the $k$-th visual layer and $\widetilde{\mathbf{p}}_i^{(l)}\in\mathbb{R}^{d_v}$ is the aggregated visual feature at that position for the $l$-th LLM layer. Stacking these vectors row-wise gives $\widetilde{\mathbf{P}}^{(l)}=[\widetilde{\mathbf{p}}_1^{(l)};\ldots;\widetilde{\mathbf{p}}_{N_v}^{(l)}]\in\mathbb{R}^{N_v\times d_v}$. A layer-specific projector $\rho_l:\mathbb{R}^{d_v}\to\mathbb{R}^{d}$ is then applied row-wise, $\widehat{\mathbf{Z}}^{(l)}=\rho_l(\widetilde{\mathbf{P}}^{(l)})\in\mathbb{R}^{N_v\times d}$, and the result is injected back into the visual hidden states through a gated residual connection:
\begin{equation}
\widetilde{\mathbf{H}}_v^{(l)} = \mathbf{H}_v^{(l)} + \gamma_l \widehat{\mathbf{Z}}^{(l)},
\quad \gamma_l = \sigma(a_l),
\label{eq:gated_residual}
\end{equation}
where $a_l$ is a learnable scalar and $\sigma$ is the sigmoid function; the gate $\gamma_l$ controls how strongly visual information is injected at the current layer.

Through these dynamic gated pathways, the visual hierarchy interacts with the LLM at every layer: the visual representation evolves with decoding depth rather than being committed up front by a one-shot, fixed injection.

\subsection{Training Objective}

To prevent the routing distribution from collapsing onto a small subset of visual layers, we augment the standard autoregressive loss with a lightweight \emph{entropy regularizer} that maximizes the entropy of the batch-averaged routing distribution at every LLM layer:
\begin{equation}
\mathcal{L}_{\mathrm{total}}
=
-\frac{1}{|\mathcal{M}|}
\sum_{t\in\mathcal{M}}
\log p_{\theta}\!\left(x_t\mid x_{<t}, \{\widehat{\mathbf{Z}}^{(l)}\}_{l=1}^{L}\right)
\;-\;
\lambda\,
\frac{1}{L}
\sum_{l=1}^{L}
H\!\left(\bar{\boldsymbol{\omega}}^{(l)}\right),
\quad
H(\boldsymbol{p})=-\sum_{k=1}^{L_v} p_k\log p_k.
\end{equation}
Here $\mathcal{M}$ denotes the set of valid supervision positions, $\{\widehat{\mathbf{Z}}^{(l)}\}_{l=1}^{L}$ are the per-layer fused visual representations injected into the LLM, $\bar{\boldsymbol{\omega}}^{(l)}=(\bar{\omega}_1^{(l)},\ldots,\bar{\omega}_{L_v}^{(l)})^{\top}$ is the batch-averaged routing distribution at LLM layer $l$, $H(\cdot)$ is the Shannon entropy, and $\lambda\geq 0$ controls the regularization strength. The second term rewards higher entropy and therefore discourages collapse onto a few visual layers. Following TGIF~\cite{lin2026text}, we set $\lambda=0.01$ during pre-training and $\lambda=0$ during instruction fine-tuning so that routing can specialize to task-relevant visual granularities.

\section{Experiments}

\subsection{Experimental Settings}

\textbf{Implementation Details and Training Recipe.}
We follow the LLaVA-1.5~\cite{liu2024improved} training recipe, using CLIP-ViT-L/14-336~\cite{radford2021learning} as the vision encoder and Vicuna as the LLM backbone. Training proceeds in two stages: pre-training (PT) on LLaVA-558K~\cite{liu2024improved} and supervised fine-tuning (SFT) on LLaVA-mixed-665K~\cite{liu2024improved}. Because \name interacts with the hidden states at every LLM layer, both stages jointly optimize the LLM together with the newly introduced \name modules. The vision encoder is kept frozen throughout, and each stage is trained for one epoch.

All hyperparameters follow LLaVA-1.5: AdamW with cosine scheduling (warmup ratio $0.03$), batch sizes of $256/128$ and learning rates of $1\times10^{-3}/2\times10^{-5}$ for PT/SFT, respectively. The routing MLP uses a hidden dimension of $1024$, and the residual gate is initialized to $0.3$. Following TGIF~\cite{lin2026text}, the entropy regularizer uses $\lambda=0.01$ during PT and is disabled ($\lambda=0$) during SFT so routing can specialize. More hyperparameter sensitivity results are provided in Appendix~\ref{sec:appendix_hyperparam}.

\textbf{Benchmarks.}
We evaluate on two complementary suites of multimodal benchmarks. \emph{Academic VQA} covers GQA~\cite{hudson2019gqa}, SciQA-IMG~\cite{lu2022learn}, VQAv2~\cite{goyal2017making}, and VizWiz~\cite{gurari2018vizwiz}. \emph{Instruction-following / LMM benchmarks} cover SEED-Bench (SEED/SEED$^{I}$)~\cite{li2023seed}, LLaVA-Wild~\cite{liu2023visual}, MM-Vet~\cite{yu2023mm}, MMBench~\cite{liu2024mmbench}, and POPE~\cite{li2023evaluating}, jointly probing multimodal comprehension, fine-grained perception, and hallucination resistance.

\textbf{Baselines.}
We compare \name against two groups of baselines. The first group consists of general-purpose MLLMs: BLIP-2~\cite{li2023blip}, InstructBLIP~\cite{dai2023instructblip}, Shikra~\cite{chen2023shikra}, IDEFICS~\cite{laurencon2023idefics}, Qwen-VL~\cite{bai2023qwen}, and LLaVA-1.5~\cite{liu2024improved}. The second group focuses on \emph{multi-layer visual feature utilization}: MMFuser~\cite{cao2024mmfuser}, Dense Connector~\cite{yao2024dense}, DeepStack-L~\cite{meng2024deepstack}, Sparse Shortcuts~\cite{zhang2026sparse}, IGVA~\cite{li2026instruction}, and TGIF~\cite{lin2026text}. For a fair comparison, we focus on methods that share the Vicuna backbone and the LLaVA-1.5 configuration.

\begin{table*}[t]
\centering
\caption{\textbf{Comparison with state-of-the-art MLLMs.} We evaluate on Academic-oriented VQA tasks and recent LMM benchmarks. Best results are in \textbf{bold}; second-best are \underline{underlined}. \textquotedblleft\texttt{--}\textquotedblright{} denotes results not reported in the original papers. Rows below \protect\name{} report the absolute point gain ($\uparrow$) or loss ($\downarrow$) relative to LLaVA-1.5.}
\label{tab:main_comparison}
\vspace{0.2cm}
\setlength{\tabcolsep}{0.4pt}
\renewcommand{\arraystretch}{1.02}
\small
\resizebox{\textwidth}{!}{
\begin{tabular}{@{}cc|cccc|ccccccc@{}}
\toprule
\multicolumn{2}{c|}{}
& \multicolumn{4}{c|}{\textbf{\textit{Academic-oriented VQA}}}
& \multicolumn{7}{c}{\textbf{\textit{LMM benchmarks}}} \\
\textbf{Method} & \textbf{LLM}
& \textbf{GQA} & \textbf{SQA$^{I}$} & \textbf{VQA$^{v2}$} & \textbf{VizWiz}
& \textbf{SEED} & \textbf{SEED$^{I}$} & \textbf{LLaVA$^{W}$} & \textbf{MM-Vet} & \textbf{MMB} & \textbf{MMB$^{CN}$} & \textbf{POPE} \\
\midrule
BLIP-2~\cite{li2023blip} & Vicuna-13B & 41.0 & 61.0 & 65.0 & 19.6 & 46.4 & 49.7 & 38.1 & 22.4 & -- & -- & 85.3 \\
InstructBLIP~\cite{dai2023instructblip} & Vicuna-7B & 49.2 & 60.5 & -- & 34.5 & 53.4 & 58.8 & 60.9 & 26.2 & 36.0 & 23.7 & -- \\
InstructBLIP~\cite{dai2023instructblip} & Vicuna-13B & 49.5 & 63.1 & -- & 33.4 & 78.9 & -- & 58.2 & 25.6 & -- & -- & 78.9 \\
Shikra~\cite{chen2023shikra} & Vicuna-13B & -- & -- & 77.4 & -- & -- & -- & -- & -- & 58.8 & -- & -- \\
IDEFICS-9B~\cite{laurencon2023idefics} & LLaMA-7B & 38.4 & -- & 50.9 & 35.5 & -- & 44.5 & -- & -- & 48.2 & 25.2 & -- \\
IDEFICS-80B~\cite{laurencon2023idefics} & LLaMA-65B & 45.2 & -- & 60.0 & 36.0 & -- & 53.2 & -- & -- & 54.5 & 38.1 & -- \\
Qwen-VL~\cite{bai2023qwen} & Qwen-7B & 59.3 & 67.1 & 78.8 & 35.2 & 56.3 & 62.3 & -- & -- & 38.2 & 7.4 & -- \\
Qwen-VL-Chat~\cite{bai2023qwen} & Qwen-7B & 57.5 & 68.2 & 78.2 & 38.9 & 58.2 & 65.4 & -- & -- & 60.6 & 56.7 & -- \\
\midrule
\rowcolor{sectionbg} LLaVA-1.5~\cite{liu2024improved} & Vicuna-7B & 62.0 & 66.8 & 78.5 & 50.0 & 58.6 & 66.1 & 63.4 & 30.5 & 64.3 & 58.3 & 85.9 \\
MMFuser~\cite{cao2024mmfuser} & Vicuna-7B & 62.8 & 68.7 & 79.1 & 53.4 & \underline{60.8} & -- & \underline{65.5} & 32.6 & \textbf{67.5} & \underline{60.1} & 86.3 \\
Dense Connector~\cite{yao2024dense} & Vicuna-7B & \underline{63.8} & 69.5 & 79.5 & -- & -- & -- & -- & 32.7 & 66.8 & -- & 86.6 \\
DeepStack-L~\cite{meng2024deepstack} & Vicuna-7B & 63.1 & -- & 79.5 & -- & 60.6 & -- & -- & 29.9 & -- & -- & 86.7 \\
SparseCut~\cite{zhang2026sparse} & Vicuna-7B & 63.6 & 70.0 & \textbf{80.0} & \underline{54.2} & -- & 67.3 & -- & -- & \underline{67.3} & \underline{60.1} & \underline{87.7} \\
IGVA~\cite{li2026instruction} & Vicuna-7B & 63.1 & \underline{70.2} & -- & \textbf{54.3} & -- & \textbf{68.3} & -- & \underline{33.5} & 66.9 & -- & \textbf{87.8} \\
TGIF~\cite{lin2026text} & Vicuna-7B & 62.6 & 70.1 & -- & -- & -- & -- & -- & -- & 66.4 & -- & 86.2 \\
\rowcolor[RGB]{224,255,255} \name & Vicuna-7B & \textbf{64.6} & \textbf{71.0} & \underline{79.7} & 53.1 & \textbf{61.5} & \underline{67.5} & \textbf{66.5} & \textbf{33.8} & 66.7 & \textbf{60.5} & 87.0 \\
& & \textcolor{gray}{$\uparrow$2.6} & \textcolor{gray}{$\uparrow$4.2} & \textcolor{gray}{$\uparrow$1.2} & \textcolor{gray}{$\uparrow$3.1} & \textcolor{gray}{$\uparrow$2.9} & \textcolor{gray}{$\uparrow$1.4} & \textcolor{gray}{$\uparrow$3.1} & \textcolor{gray}{$\uparrow$3.3} & \textcolor{gray}{$\uparrow$2.4} & \textcolor{gray}{$\uparrow$2.2} & \textcolor{gray}{$\uparrow$1.1} \\
\midrule
\rowcolor{sectionbg} LLaVA-1.5~\cite{liu2024improved} & Vicuna-13B & 63.3 & 71.6 & 80.0 & 53.6 & 61.6 & 68.2 & \underline{70.7} & 35.4 & 67.7 & 63.6 & 85.9 \\
MMFuser~\cite{cao2024mmfuser} & Vicuna-13B & 63.4 & 71.2 & 80.1 & \textbf{57.4} & 62.0 & -- & \textbf{71.8} & \underline{36.6} & \underline{69.9} & 63.8 & 87.5 \\
DeepStack-L~\cite{meng2024deepstack} & Vicuna-13B & \underline{64.2} & -- & \underline{80.9} & -- & \textbf{63.5} & -- & -- & 35.9 & -- & -- & \underline{87.7} \\
SparseCut~\cite{zhang2026sparse} & Vicuna-13B & 64.0 & \underline{71.8} & \textbf{81.8} & \underline{56.1} & -- & \textbf{69.0} & -- & -- & 69.7 & \underline{64.1} & \textbf{87.8} \\
\rowcolor[RGB]{224,255,255} \name & Vicuna-13B & \textbf{65.5} & \textbf{73.0} & 80.7 & \textbf{57.4} & \underline{63.0} & \textbf{69.0} & 70.0 & \textbf{37.5} & \textbf{70.1} & \textbf{65.4} & 87.6 \\
& & \textcolor{gray}{$\uparrow$2.2} & \textcolor{gray}{$\uparrow$1.4} & \textcolor{gray}{$\uparrow$0.7} & \textcolor{gray}{$\uparrow$3.8} & \textcolor{gray}{$\uparrow$1.4} & \textcolor{gray}{$\uparrow$0.8} & \textcolor{gray}{$\downarrow$0.7} & \textcolor{gray}{$\uparrow$2.1} & \textcolor{gray}{$\uparrow$2.4} & \textcolor{gray}{$\uparrow$1.8} & \textcolor{gray}{$\uparrow$1.7} \\
\bottomrule
\end{tabular}
}
\end{table*}

\subsection{Main Results}

\textbf{Comparison with SOTA Methods.}
Table~\ref{tab:main_comparison} compares \name with mainstream MLLMs and multi-layer visual feature utilization methods under Vicuna-7B and 13B settings. Taking Vicuna-7B as an illustrative example, \name consistently improves over the LLaVA-1.5 baseline---most prominently on VizWiz (+3.1), SQA$^{I}$ (+4.2), LLaVA$^{W}$ (+3.1), and MM-Vet (+3.3)---with similar trends on the remaining benchmarks. Against the strongest multi-layer baselines, \name attains the best score on GQA, SQA$^{I}$, SEED, LLaVA$^{W}$, MM-Vet, and MMB$^{CN}$, suggesting that compressing the visual hierarchy on the encoder side or using predefined sparse or hierarchical connections is insufficient to capture the full interaction between hierarchical visual evidence and the layer-wise LLM state. Overall, the gains support our central claim: dynamically gating \emph{where to connect} between ViT and LLM layers is a more effective use of the visual hierarchy than any fixed wiring.

\newsavebox{\generalizationbox}
\sbox{\generalizationbox}{%
\fontsize{8}{9}\selectfont
\setlength{\tabcolsep}{3pt}%
\renewcommand{\arraystretch}{1.0}%
\begin{tabular}{@{}lcc|cccc@{}}
\toprule
\textbf{Method} & \textbf{ViT} & \textbf{LLM} & \textbf{GQA} & \textbf{SQA$^{I}$} & \textbf{MMB} & \textbf{MM-Vet} \\
\midrule
Qwen3-VL-2B~\cite{bai2025qwen3} & SigLIP2-L & Qwen3-1.7B & 63.9 & 85.3 & 76.4 & 49.8 \\
+ \name & SigLIP2-L & Qwen3-1.7B & \textbf{65.4} & \textbf{86.9} & \textbf{77.4} & \textbf{50.9} \\
\midrule
LLaVA-NeXT~\cite{liu2024llavanext} & CLIP-L & Vicuna-7B & 64.0 & 69.5 & 66.5 & 33.1 \\
+ \name & CLIP-L & Vicuna-7B & \textbf{64.8} & \textbf{72.3} & \textbf{68.7} & \textbf{34.6} \\
\bottomrule
\end{tabular}%
}
\edef\genintextsep{\the\intextsep}%
\setlength{\intextsep}{9pt}%
\begin{wraptable}{r}{\wd\generalizationbox}
\vspace{-0.22in}
\centering
\caption{\textbf{Generalization across MLLM frameworks.} We evaluate \name on different vision encoders (SigLIP2 vs.\ CLIP), LLM backbones (Qwen3 vs.\ Vicuna), and high-resolution tiling.}
\label{tab:generalization}
\vspace{0.2cm}
\usebox{\generalizationbox}%
\end{wraptable}
\setlength{\intextsep}{\genintextsep}%
\noindent\textbf{Generalization Across MLLM Frameworks.}
To verify that \name captures a general architectural principle, we evaluate it on two additional MLLM configurations that differ substantially in vision encoder, LLM backbone, and resolution strategy. For fair comparison, all models are trained on the same 558K (PT) + 665K (SFT) data splits following DenseConnector~\cite{yao2024dense}, with one epoch per stage. As shown in Table~\ref{tab:generalization}, \name delivers consistent improvements on both Qwen3-VL-2B (SigLIP2 + Qwen3-1.7B) and LLaVA-NeXT (AnyRes tiling), confirming that the \emph{where to connect} principle is architecture-agnostic.

\begin{figure*}[t]
  \centering
  \includegraphics[width=1\linewidth,height=0.9\textheight,keepaspectratio]{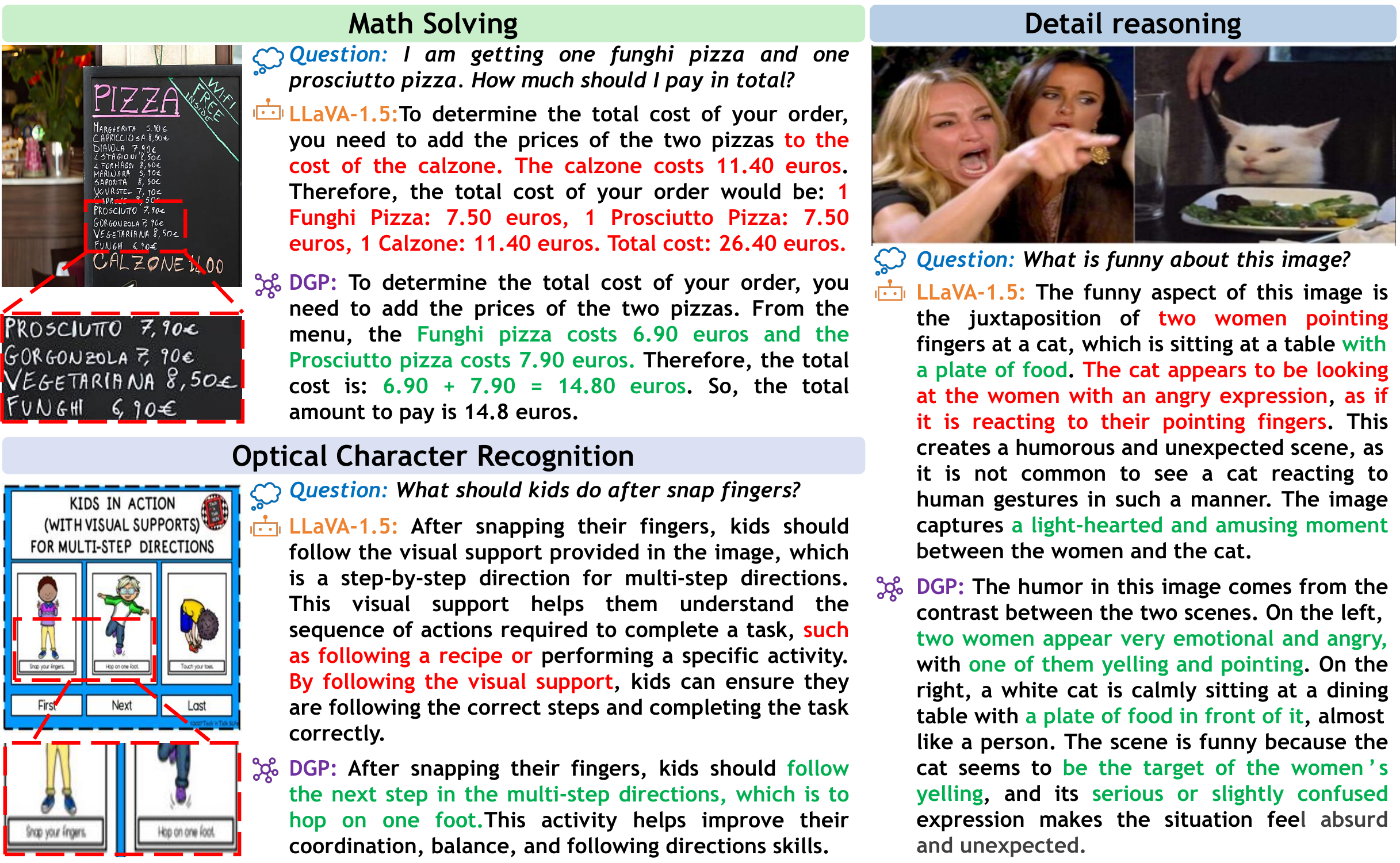}
  \vspace{-0.15in}
  \caption{\textbf{Qualitative comparison between \name and LLaVA-1.5.} Representative examples from math solving, OCR understanding, and detailed visual reasoning. LLaVA-1.5 often misses critical visual evidence or hallucinates content; \name produces more accurate, evidence-grounded responses by adaptively accessing multi-granularity visual features.}
  \label{fig:case_study}
  \vspace{-0.06in}
\end{figure*}

\textbf{Case Study.}
To make the effect of \name more concrete, Figure~\ref{fig:case_study} contrasts \name with LLaVA-1.5 on math reasoning, OCR understanding, and fine-grained visual reasoning. LLaVA-1.5 frequently fails to read critical visual evidence or forms wrong semantic associations---for example, misreading menu prices or hallucinating non-existent items such as a ``calzone''---and this propagates into clearly hallucinated reasoning. \name, in contrast, uses its dynamic gated pathways to let different LLM layers access multi-granularity visual cues on demand, accurately recovering textual and numerical details, facial expressions, and other fine-grained evidence, and grounds the final answer in those cues.

\newsavebox{\efficiencybox}
\sbox{\efficiencybox}{%
\fontsize{8}{9}\selectfont
\setlength{\tabcolsep}{3pt}%
\renewcommand{\arraystretch}{1.0}%
\begin{tabular}{@{}l r@{\,}l r@{\,}l r@{\,}l @{}}
\toprule
\textbf{Method} & \multicolumn{2}{c}{\textbf{Params (M)}} & \multicolumn{2}{c}{\textbf{FLOPs (G)}} & \multicolumn{2}{c}{\textbf{Mem (GB)}} \\
\midrule
LLaVA-1.5-7B & 6{,}759.4 & & 9{,}168.3 & & 15.51 & \\
+ \name & 7{,}117.3 & \textcolor{gray}{\scriptsize +5.29\%} & 9{,}541.2 & \textcolor{gray}{\scriptsize +4.07\%} & 15.49 & \textcolor{gray}{\scriptsize $-$0.13\%} \\
\midrule
LLaVA-1.5-13B & 13{,}047.3 & & 17{,}731.2 & & 27.34 & \\
+ \name & 13{,}604.9 & \textcolor{gray}{\scriptsize +4.27\%} & 17{,}718.8 & \textcolor{gray}{\scriptsize $-$0.07\%} & 28.47 & \textcolor{gray}{\scriptsize +4.13\%} \\
\bottomrule
\end{tabular}%
}
\edef\effintextsep{\the\intextsep}%
\setlength{\intextsep}{9pt}%
\begin{wraptable}{r}{\wd\efficiencybox}
\vspace{-0.22in}
\centering
\caption{Efficiency analysis of \name at batch size 1. FLOPs and peak memory are measured during end-to-end autoregressive generation.}
\label{tab:efficiency_main}
\vspace{0.2cm}
\usebox{\efficiencybox}%
\end{wraptable}
\setlength{\intextsep}{\effintextsep}%
\noindent\textbf{Efficiency Analysis.}
Table~\ref{tab:efficiency_main} reports the computational overhead of \name. The DGP module introduces only 5.29\% additional parameters for the 7B model and 4.27\% for the 13B model. The FLOPs overhead is modest for 7B (+4.07\%) and negligible for 13B ($-$0.07\%), while peak memory is nearly unchanged for 7B ($-$0.13\%) and increases by 4.13\% for 13B. Taken together with the performance gains, these modest overheads indicate a favorable trade-off between multimodal performance and computational cost.

\subsection{Ablation Studies and Analysis}

\textbf{Ablation of \name Components.}
Table~\ref{tab:ablation} ablates the two key ingredients of \name---gated routing and entropy regularization---and keeps a uniform average over the 24 ViT layers as a control, so as to disentangle gains coming from ``using more visual layers'' from those coming from \emph{where to connect}. Adding \emph{\textbf{gated routing alone}} already yields consistent improvements over the LLaVA-1.5 baseline on most benchmarks. By contrast, \emph{\textbf{uniform average fusion}} without gated routing hurts performance on most benchmarks relative to the baseline. Equal-weight mixing across the visual hierarchy appears to blend incompatible granularities and weaken fine-grained evidence, whereas gated routing keeps connectivity selective and conditioned on the language-layer state. Stacking the \emph{\textbf{entropy regularizer}} on top of gated routing brings a further gain, indicating that a well-spread routing distribution prevents collapse onto a few visual layers and stabilizes training. Together, these results show that the improvement is driven by the \emph{where to connect} mechanism---routing that adapts to the language-layer state---rather than by the mere availability of more visual features.

\begin{table*}[t]
  \centering
  \caption{\textbf{Ablation of \name components.} We evaluate the contribution of average fusion, gated routing, and entropy regularization.}
  \label{tab:ablation}
\vspace{0.2cm}
  \setlength{\tabcolsep}{1.1pt}
  \renewcommand{\arraystretch}{1.02}
  \small
  \resizebox{\textwidth}{!}{
  \begin{tabular}{@{}ccc|cccc|ccccccc@{}}
  \toprule
  \textbf{Avg.} & \textbf{Gated} & \textbf{Entropy}
  & \multicolumn{4}{c|}{\textbf{\textit{Academic-oriented VQA}}}
  & \multicolumn{7}{c}{\textbf{\textit{LMM benchmarks}}} \\
  \textbf{Fusion} & \textbf{Routing} & \textbf{Reg.}
  & \textbf{GQA} & \textbf{SQA$^{I}$} & \textbf{VQA$^{v2}$} & \textbf{VizWiz}
  & \textbf{SEED} & \textbf{SEED$^{I}$} & \textbf{LLaVA$^{W}$} & \textbf{MM-Vet} & \textbf{MMB} & \textbf{MMB$^{CN}$} & \textbf{POPE} \\
  \midrule
  {\normalfont\xmark} & {\normalfont\xmark} & {\normalfont\xmark}
  & 62.0 & 66.8 & 78.5 & 50.0
  & 58.6 & 66.1 & 63.4 & 30.5 & 64.3 & 58.3 & 85.9 \\
  
  {\normalfont\cmark} & {\normalfont\xmark} & {\normalfont\xmark}
  & 60.2 & 64.3 & 72.0 & 50.2
  & 57.9 & 63.3 & 56.3 & 25.3 & 57.6 & 44.9 & 77.2 \\
  
  {\normalfont\xmark} & {\normalfont\cmark} & {\normalfont\xmark}
  & 64.1 & 70.5 & 78.9 & 52.0
  & 60.9 & 66.7 & 63.6 & 32.4 & 65.6 & 59.1 & 86.4 \\
  
  {\normalfont\xmark} & {\normalfont\cmark} & {\normalfont\cmark}
  & \textbf{64.6} & \textbf{71.0} & \textbf{79.7} & \textbf{53.1}
  & \textbf{61.5} & \textbf{67.5} & \textbf{66.5} & \textbf{33.8} & \textbf{66.7} & \textbf{60.5} & \textbf{87.0} \\
  \bottomrule
  \end{tabular}
  }
  \end{table*}
  
  \begin{figure*}[t]
    \centering
    \includegraphics[width=1\linewidth]{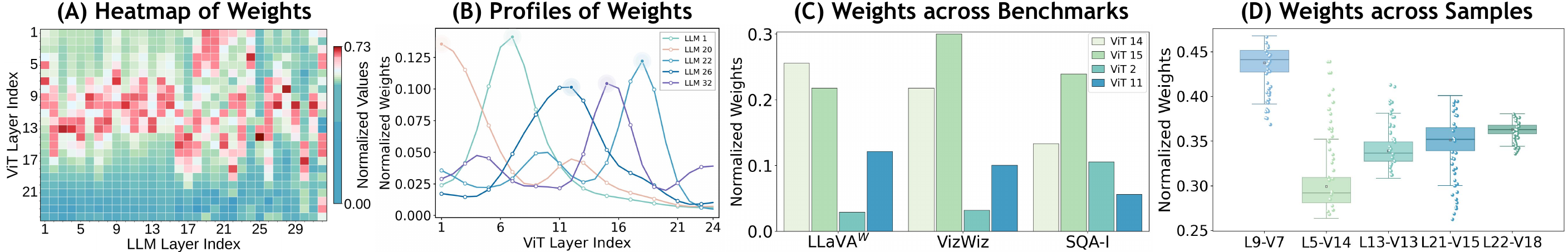}
    \vspace{-0.15in}
    \caption{\textbf{Gated Pathway Analysis.} (A) Heatmap of routing weights over LLM (x) and ViT (y) layers. (B) Routing curves of representative LLM layers with multimodal patterns. (C) Routing weights of LLM layer~21 across benchmarks, showing task-specific preferences. (D) Per-sample routing distributions within the same benchmark, showing sample-level diversity.}
    \label{fig:routing_analysis}
    \vspace{-0.08in}
  \end{figure*}
  
  \begin{figure*}[t]
    \centering
    \includegraphics[width=1\linewidth]{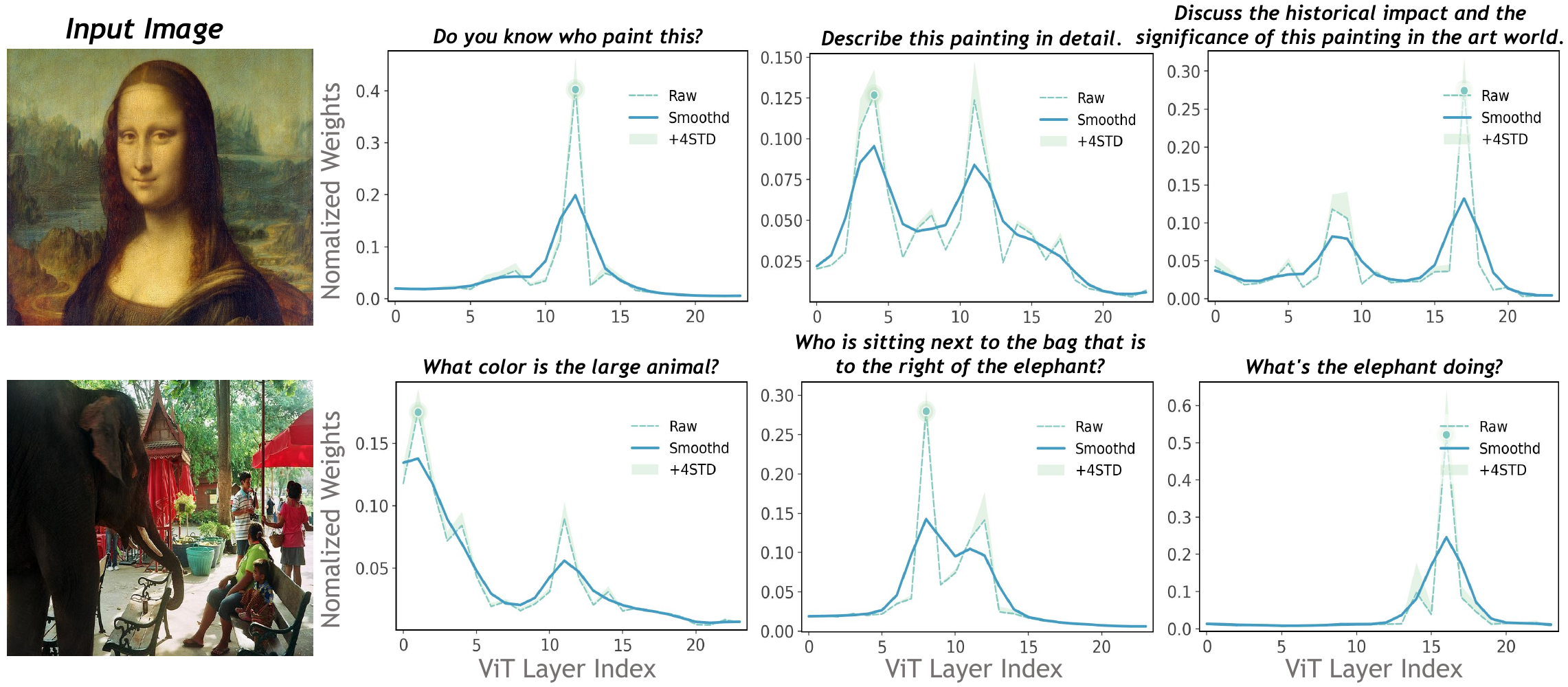}
    \vspace{-0.15in}
    \caption{\textbf{Task-aware routing for the same image.} As the question changes, routing shifts: attribute recognition activates shallower layers, while global-semantic questions activate deeper ones.}
    \label{fig:task_routing}
    \vspace{-0.06in}
  \end{figure*}
  
  \begin{figure*}[t]
    \centering
    \includegraphics[width=1\linewidth]{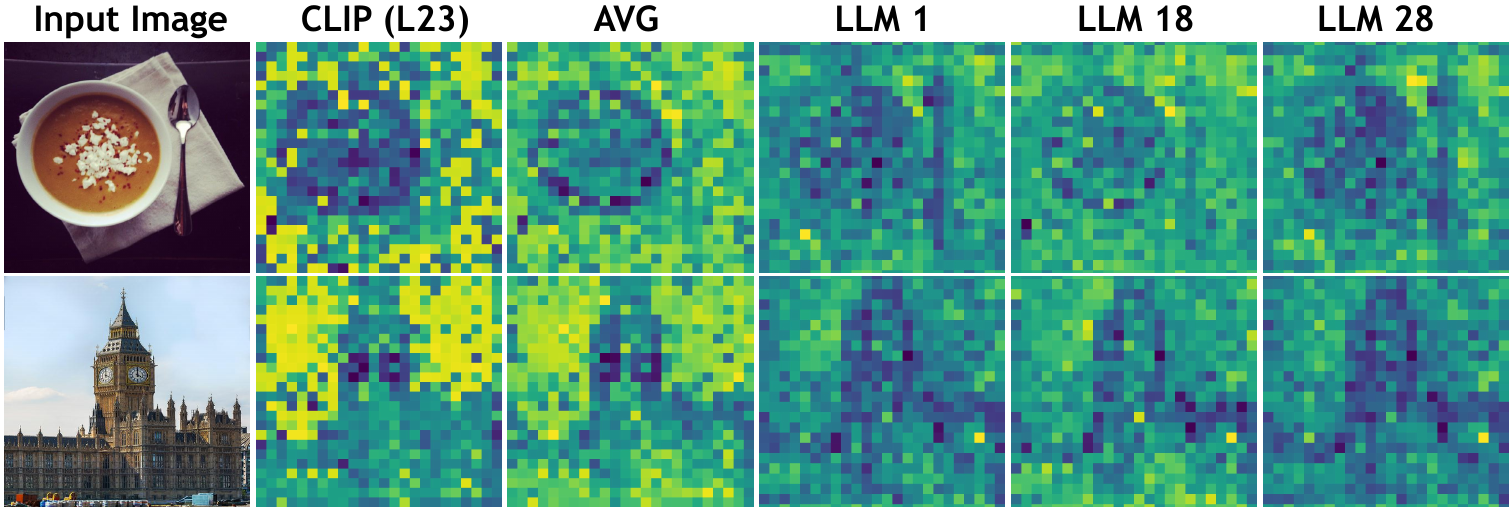}
    \vspace{-0.15in}
    \caption{\textbf{Fused visual feature visualization.} We compare CLIP layer~23, the uniform average over all 24 ViT layers, and \name's gated fusion at different LLM layers.}
    \label{fig:feature_vis}
    \vspace{-0.06in}
  \end{figure*}

\textbf{Layer-wise Gated Pathway Analysis.}
To inspect what \name actually learns, we visualize the routing weights of every LLM layer over the ViT layers. Figure~\ref{fig:routing_analysis}(A) shows a clearly non-uniform, layer-dependent distribution: \name neither fixates on the last ViT layer nor degenerates into a uniform average. Middle ViT layers are activated most frequently, suggesting that they offer a useful trade-off between local structure and high-level semantics. Shallow and deep features are also selected by certain LLM layers, confirming that the model dynamically reaches into different visual granularities according to where it is in decoding rather than relying on a single visual layer.

Figure~\ref{fig:routing_analysis}(B) zooms in on representative LLM layers. Their routing curves peak at different positions: LLM layer~1 favors shallow-to-middle ViT layers, LLM layer~20 leans toward shallow features, whereas LLM layers~22 and~32 peak at mid-deep and deep ViT layers, respectively. The curves are typically multimodal rather than unimodal or flat, showing that \name selectively combines several visual layers per LLM layer and that different LLM layers prefer different visual granularities depending on their language state. These visualizations confirm that \name discovers non-uniform, dynamic ViT--LLM connection patterns, exploiting the complementarity of the visual hierarchy.

\textbf{Task-aware Gated Pathway Analysis.}
We next ask whether the routing also adapts to the task itself. Figure~\ref{fig:routing_analysis}(C) plots the routing weights of LLM layer~21 across benchmarks: each benchmark induces a distinct visual-layer profile. LLaVA$^{W}$ and VizWiz allocate more mass to mid-level features, in line with the need to balance object-level semantics and local visual cues for open-ended QA and real-world understanding, whereas SQA$^{I}$ additionally puts noticeable weight on shallow ViT layer~2, reflecting the fine-grained details required in scientific reasoning. Figure~\ref{fig:routing_analysis}(D) zooms one level further down: even within a single benchmark, individual samples exhibit substantial routing diversity, indicating that \name's adaptivity goes beyond benchmark-level differences and operates at the sample level.

Figure~\ref{fig:task_routing} fixes the image and varies the question. For questions about attribute recognition or small-region details, \name activates shallow or shallow-to-middle ViT features---layers that preserve edges, textures and local spatial structure. For questions that rely more on global semantics or abstract reasoning---painting identification, historical significance, action understanding---the routing weight peaks instead at mid-deep or deep ViT layers. For open-ended questions such as ``describe the image in detail,'' which require both local details and holistic semantics, the routing distribution naturally widens into a broader multimodal shape that simultaneously taps multiple visual hierarchies. These patterns indicate that \name's routing is dynamically conditioned on the layer-wise linguistic context and semantic state of the LLM, rather than being determined solely by fixed visual-layer priors.

\textbf{Feature Visualization.}
Figure~\ref{fig:feature_vis} compares fixed single-layer CLIP features (L23), a uniform average over all 24 ViT layers, and the features produced by \name at different LLM layers. CLIP (L23) yields coarse spatial responses with noisy activations in the background; naive average fusion brings in multi-layer information but, because features at different semantic granularities are superposed without selection, still suffers from diffuse responses and amplified local noise. \name's LLM-layer-specific features instead show clearly more stable spatial structure, sharper subject contours, and lower background noise, preserving local details such as spoon edges and building boundaries. Crucially, the fused features differ across LLM layers~1, 18, and 28, confirming that \name generates differentiated visual representations via dynamic routing rather than collapsing into a fixed averaging pattern. Overall, \name strengthens global semantics while keeping shallow and middle-layer cues intact, mitigating both the information bottleneck of fixed single-layer features and the noise accumulation of naive averaging.

\section{Conclusion}


In this work, we reframe visual information injection in MLLMs around the question of \emph{where to connect}, and argue that fixed single-layer inputs or predefined sparse or hierarchical connections cannot match the differentiated demands of hierarchical ViT representations and layer-wise LLM semantics. We answer this question with \name (Dynamic Gated Pathways), an all-to-all cross-layer connection mechanism in which each LLM layer adaptively gates its preference over every ViT layer based on its current semantic state, integrating multi-granularity visual information on demand at every decoding depth. \name delivers consistent gains across a broad set of multimodal benchmarks, while routing analyses and feature visualizations show that it learns interpretable, layer-dependent connection patterns that explain where the gains come from.


\begin{ack}
This work was supported by the National Natural Science Foundation of China (Nos.~62472359 and 62372379).
\end{ack}

\end{CJK*}


{
\small
\bibliographystyle{plainnat}
\bibliography{main}

}

\newpage

\appendix

\begin{CJK*}{UTF8}{gbsn}

\section{Methodological Details}
\label{sec:appendix_method}

Algorithm~\ref{alg:dgp} summarizes the overall procedure of \name. Table~\ref{tab:token_flow} details the token flow and tensor shapes of \name under the Vicuna-7B setting.

\subsection{Algorithm}

\begin{algorithm}[H]
\caption{Dynamic Gated Pathways (\name)}
\label{alg:dgp}
\begin{algorithmic}[1]
\Require Input image $\mathbf{I} \in \mathbb{R}^{H \times W \times 3}$, text tokens $\mathbf{T} \in \mathbb{R}^{N_t \times d}$
\Require Vision encoder $F(\cdot)$ with $L_v$ layers, LLM with $L$ layers
\Require Shared cross-attention $\mathrm{MHA}_{\mathrm{shared}}$, layer-specific routing MLPs $\{\mathrm{MLP}^{(l)}\}_{l=1}^{L}$
\Require Layer-specific projectors $\{\rho_l\}_{l=1}^{L}$, gate scalars $\{a_l\}_{l=1}^{L}$
\Require Learnable glimpse token $\mathbf{g} \in \mathbb{R}^{d}$, layer ID embeddings $\mathbf{E}_{\mathrm{id}} \in \mathbb{R}^{L_v \times d}$

\Statex \textit{\textcolor[RGB]{50,134,187}{// Step 1: Vision Encoding and Feature Caching}}
\State $\{\mathbf{X}_v^{(\ell)}\}_{\ell=1}^{L_v} \leftarrow F(\mathbf{I})$ \Comment{Extract hierarchical visual features}
\For{$\ell = 1, \dots, L_v$}
    \State $\mathbf{P}^{(\ell)} \leftarrow \mathbf{X}_v^{(\ell)}[:, 1:]$ \Comment{Patch features: $[B, N_v, d_v]$}
    \State $\mathbf{c}^{(\ell)} \leftarrow \mathbf{X}_v^{(\ell)}[:, 0]$ \Comment{CLS token: $[B, d_v]$}
    \State $\widetilde{\mathbf{c}}^{(\ell)} \leftarrow \mathrm{Linear}(\mathrm{LN}(\mathbf{c}^{(\ell)}))$ \Comment{Project CLS to LLM space: $[B, d]$}
\EndFor
\State $\mathbf{K} = \mathbf{V} \leftarrow \mathrm{Stack}(\widetilde{\mathbf{c}}^{(1)}, \dots, \widetilde{\mathbf{c}}^{(L_v)}) + \mathbf{E}_{\mathrm{id}}$ \Comment{CLS summaries with layer IDs: $[B, L_v, d]$}

\Statex \textit{\textcolor[RGB]{50,134,187}{// Step 2: Initial Visual Token Injection}}
\State $\mathbf{V}_0 \leftarrow \mathrm{mm\_projector}(\mathbf{P}^{(L_v-1)})$ \Comment{Penultimate-layer patches to LLM space: $[B, N_v, d]$}
\State $\mathbf{H}^{(0)} \leftarrow \mathrm{Concat}(\mathbf{V}_0, \mathbf{T}, \mathbf{g})$ \Comment{Multimodal input sequence}

\Statex \textit{\textcolor[RGB]{50,134,187}{// Step 3: Layer-wise Dynamic Gated Pathway Injection}}
\For{$l = 1, \dots, L$}
    \State $\mathbf{h}_g^{(l)} \leftarrow \mathbf{H}^{(l-1)}[\mathrm{glimpse\_pos}]$ \Comment{Glimpse hidden state: $[B, 1, d]$}
    \State $\mathbf{Q}^{(l)} \leftarrow \mathbf{h}_g^{(l)}$ \Comment{Routing query (as in Methodology)}
    \State $\mathbf{r}^{(l)} \leftarrow \mathrm{MHA}_{\mathrm{shared}}(\mathbf{Q}^{(l)}, \mathbf{K}, \mathbf{V}) + \mathbf{Q}^{(l)}$ \Comment{Cross-attention with residual}
    \State $\boldsymbol{\omega}^{(l)} \leftarrow \mathrm{Softmax}(\mathrm{MLP}^{(l)}(\mathrm{LN}(\mathbf{r}^{(l)})))$ \Comment{Routing weights: $[B, L_v]$}
    \State $\widetilde{\mathbf{P}}^{(l)} \leftarrow \sum_{k=1}^{L_v} \omega_k^{(l)} \cdot \mathbf{P}^{(k)}$ \Comment{Weighted patch fusion: $[B, N_v, d_v]$}
    \State $\widehat{\mathbf{Z}}^{(l)} \leftarrow \rho_l(\widetilde{\mathbf{P}}^{(l)})$ \Comment{Project to LLM space: $[B, N_v, d]$}
    \State $\gamma_l \leftarrow \sigma(a_l)$ \Comment{Gating coefficient}
    \State $\widetilde{\mathbf{H}}_v^{(l-1)} \leftarrow \mathbf{H}_v^{(l-1)} + \gamma_l \cdot \widehat{\mathbf{Z}}^{(l)}$ \Comment{Gated residual injection (Methodology Eq.~\eqref{eq:gated_residual})}
    \State $\mathbf{H}^{(l)} \leftarrow \mathrm{LLMLayer}_l(\mathbf{H}^{(l-1)} \text{ with } \mathbf{H}_v^{(l-1)} \leftarrow \widetilde{\mathbf{H}}_v^{(l-1)})$ \Comment{Forward with injected visual segment}
\EndFor

\State \textbf{Return} Final hidden states $\mathbf{H}^{(L)}$ for autoregressive decoding
\end{algorithmic}
\end{algorithm}

\subsection{Token Flow in \name (Vicuna-7B Example)}

Taking Vicuna-7B (32 LLM layers, hidden dimension 4096) and CLIP-ViT-L/14-336 (24 ViT layers, hidden dimension 1024, $24 \times 24 = 576$ patches) as an example, Table~\ref{tab:token_flow} details the token flow and dimension changes at each stage of \name.

\begin{table*}[!htbp]
\centering
\caption{\textbf{Token flow and tensor dimensions in \name.} We illustrate each processing stage using Vicuna-7B (32 layers, $d=4096$) and CLIP-ViT-L/14-336 (24 layers, $d_v=1024$, $N_v=576$ patches). $B$ denotes batch size, $N_t$ denotes the number of text tokens.}
\label{tab:token_flow}
\vspace{0.2cm}
\begingroup
\setlength{\tabcolsep}{3pt}%
\renewcommand{\arraystretch}{1.12}%
\small
\resizebox{\textwidth}{!}{
\begin{tabular}{@{}l|l|c|l@{}}
\toprule
\textbf{Stage} & \textbf{Operation} & \textbf{Tensor Shape} & \textbf{Description} \\
\midrule
\multicolumn{4}{@{}l}{\textit{\textbf{Vision Encoding (ViT)}}} \\
\midrule
Raw patches & ViT layer $\ell$ output & $[B, 577, 1024]$ & 1 CLS + 576 patches per layer \\
Patch features & Remove CLS & $[B, 576, 1024] \times 24$ & Cache all 24 layers for routing \\
CLS tokens & Extract CLS & $[B, 1, 1024] \times 24$ & Layer summaries for cross-attention \\
\midrule
\multicolumn{4}{@{}l}{\textit{\textbf{CLS Projection for Routing}}} \\
\midrule
CLS projection & $\mathrm{LN}(1024) \to \mathrm{Linear}(1024 \to 4096)$ & $[B, 1, 4096] \times 24$ & Project each CLS to LLM space \\
Layer-ID keys/values & $\mathrm{Stack}(\widetilde{\mathbf{c}}^{(1)}{+}\mathbf{e}_1,\ldots,\widetilde{\mathbf{c}}^{(L_v)}{+}\mathbf{e}_{L_v})$ & $[B, L_v, d]$ & $\mathbf{K},\mathbf{V}$ for $\mathrm{MHA}_{\mathrm{shared}}$ ($L_v{=}24$, $d{=}4096$) \\
\midrule
\multicolumn{4}{@{}l}{\textit{\textbf{Initial LLM Input Construction}}} \\
\midrule
Visual tokens & $\mathrm{mm\_projector}(\mathbf{P}^{(L_v-1)})$ & $[B, 576, 4096]$ & Penultimate ViT patches $\to$ LLM space ($L_v{=}24$) \\
Text tokens & Embedding lookup & $[B, N_t, 4096]$ & Standard text embedding \\
Glimpse token & Learnable + $\mathrm{Linear} \to \mathrm{RMSNorm}$ & $[B, 1, 4096]$ & Inserted before first answer token \\
\rowcolor[RGB]{240,248,255}
\textbf{Full sequence} & $[\mathbf{V}; \mathbf{T}; \mathbf{g}; \mathbf{A}]$ & $[B, 576{+}N_t{+}1{+}N_a, 4096]$ & \texttt{[v v v ... t t t ... g a a a ...]} \\
\midrule
\multicolumn{4}{@{}l}{\textit{\textbf{Per-Layer Dynamic Gated Pathways (at LLM layer $l$)}}} \\
\midrule
Glimpse query & $\mathbf{Q}^{(l)}=\mathbf{h}_g^{(l)}$ (glimpse position) & $[B, 1, 4096]$ & Routing query from current language-layer state \\
Cross-attention & $\mathrm{MHA}_{\mathrm{shared}}(\mathbf{Q}^{(l)},\mathbf{K},\mathbf{V})$ & $[B, 1, 4096]$ & Shared cross-attention over ViT CLS summaries \\
Routing feature & $\mathrm{MHA}_{\mathrm{shared}}$ output $+$ residual & $[B, 1, 4096]$ & Input to layer-specific routing MLP \\
Routing weights & $\mathrm{MLP}^{(l)} \to \mathrm{Softmax}$ & $[B, 24]$ & Per-layer distribution over ViT layers \\
\midrule
\multicolumn{4}{@{}l}{\textit{\textbf{Per-Layer Patch Fusion and Injection (at LLM layer $l$)}}} \\
\midrule
Stacked patches & Stack 24-layer patches & $[B, 24, 576, 1024]$ & All ViT layers' patch features \\
Weighted fusion & $\sum_{\ell} \omega_\ell^{(l)} \cdot \mathbf{P}^{(\ell)}$ & $[B, 576, 1024]$ & Softmax-weighted combination \\
Fusion projection & $\rho_l: \mathrm{LN}(1024) \to \mathrm{Linear}(1024 \to 4096)$ & $[B, 576, 4096]$ & Layer-specific projector \\
Gated injection & $\mathbf{H}_v^{(l)} + \gamma_l \cdot \widehat{\mathbf{Z}}^{(l)}$ & $[B, 576, 4096]$ & Update visual token hidden states \\
\bottomrule
\end{tabular}
}
\endgroup
\end{table*}


As shown in the table, the glimpse token is inserted immediately before the answer tokens (sequence format: \texttt{v...v t...t g a...a}), allowing it to attend to all visual and textual context under causal attention. At each LLM layer, the hidden state at the glimpse position queries the CLS summaries of all 24 ViT layers to produce layer-specific routing weights, which are then used to perform a weighted fusion over the original 1024-dimensional patch features. The fused features are projected to 4096 dimensions by a layer-specific projector and injected into the visual token hidden states via gated residual connections. This mechanism enables each LLM layer to adaptively select the visual granularity that best matches its current semantic state.

\section{Training Configuration and Benchmark Descriptions}
\label{sec:appendix_settings}

\subsection{Training Configuration}


Table~\ref{tab:training_config} lists the detailed training hyperparameters for the pre-training (PT) and supervised fine-tuning (SFT) stages. We follow the LLaVA-1.5 training recipe, using CLIP-ViT-L/14-336 as the vision encoder and Vicuna as the LLM backbone. The vision encoder is kept frozen throughout both stages, each trained for one epoch. The entropy regularizer is enabled during PT to stabilize routing distributions, and disabled during SFT to allow task-specific specialization. All training and efficiency measurements are conducted on NVIDIA L40-48GB GPUs.

\textbf{Backbone Selection Rationale.}
Our primary experiments are conducted under the LLaVA-1.5 framework, which remains a widely adopted open-source baseline in the research community. Specifically, we select Vicuna-v1.5 (7B/13B) as the LLM backbone and adopt the standard LLaVA-1.5 training data (LLaVA-558K for PT and LLaVA-mixed-665K for SFT). This design choice serves two purposes: (i)~\emph{reproducibility}, as all components are publicly available and well-documented; and (ii)~\emph{fair comparison} with modern multi-layer visual fusion methods (e.g., MMFuser~\cite{cao2024mmfuser}, IGVA~\cite{li2026instruction}, TGIF~\cite{lin2026text}) that likewise build upon LLaVA-1.5. By holding the backbone architecture, training data, and recipe constant, any observed performance improvement can be cleanly attributed to the proposed \name module itself, rather than to differences in the underlying LLM, vision encoder, or training corpus.

\FloatBarrier

\begin{table*}[!htbp]
\centering
\caption{\textbf{Training hyperparameters.} Configuration for pre-training (PT) on LLaVA-558K and supervised fine-tuning (SFT) on LLaVA-mixed-665K.}
\label{tab:training_config}
\vspace{0.2cm}
\small
\begin{tabular}{@{}l|c|l|c@{}}
\toprule
\multicolumn{2}{c|}{\textbf{Pre-training (PT)}} & \multicolumn{2}{c}{\textbf{Supervised Fine-tuning (SFT)}} \\
\midrule
Global batch size & 256 & Global batch size & 128 \\
Learning rate & $1\times10^{-3}$ & Learning rate & $2\times10^{-5}$ \\
Vision encoder LR & frozen & Vision encoder LR & frozen \\
LR schedule & cosine decay & LR schedule & cosine decay \\
LR warmup ratio & 0.03 & LR warmup ratio & 0.03 \\
Epochs & 1 & Epochs & 1 \\
Optimizer & AdamW & Optimizer & AdamW \\
\midrule
Routing MLP hidden dim & 1024 & Routing MLP hidden dim & 1024 \\
Residual gate init ($\gamma$) & 0.3 & Residual gate init ($\gamma$) & 0.3 \\
Entropy regularizer $\lambda$ & 0.01 & Entropy regularizer $\lambda$ & 0 \\
\bottomrule
\end{tabular}
\end{table*}

\subsection{Benchmark Descriptions}


To validate the effectiveness of \name under varying visual granularity requirements, we adopt two complementary suites of multimodal benchmarks. Different tasks exhibit significantly different demands on visual granularity: fine-grained OCR and attribute recognition rely more on shallow local features, while scene understanding and abstract reasoning require deep global semantics. This diversity in granularity requirements directly motivates the design of \name's dynamic gated pathways---enabling different LLM layers to adaptively select the visual granularity that best matches their current needs. Table~\ref{tab:benchmarks} summarizes the benchmark taxonomy and evaluated capabilities.

\FloatBarrier

\begin{table*}[!htbp]
\centering
\caption{\textbf{Benchmark descriptions.} We evaluate on two complementary suites: academic task-oriented benchmarks and instruction-following / LMM benchmarks.}
\label{tab:benchmarks}
\vspace{0.2cm}
\begingroup
\setlength{\tabcolsep}{3pt}
\renewcommand{\arraystretch}{1.06}
\footnotesize
\begin{tabularx}{\textwidth}{@{}c|l|l|X@{}}
\toprule
\textbf{Category} & \textbf{Benchmark} & \textbf{Task Type} & \textbf{Evaluated Capabilities} \\
\midrule
\multirowcell{4}{\rotatebox[origin=c]{90}{\textbf{Academic VQA}}}
& VQA$^{v2}$~\cite{goyal2017making} & General VQA & Basic image understanding, object attribute recognition, and commonsense visual reasoning \\
& GQA~\cite{hudson2019gqa} & Compositional VQA & Compositional visual reasoning with structured scene-graph annotations (relations, attributes, and logical consistency) \\
& VizWiz~\cite{gurari2018vizwiz} & Real-world VQA & Real user questions from visually impaired users with often noisy images, testing robust visual understanding \\
& SQA$^{I}$~\cite{lu2022learn} & Science QA (image) & Scientific knowledge, image-grounded reasoning, and multi-step inference \\
\midrule
\multirowcell{6}{\rotatebox[origin=c]{90}{\textbf{LMM Bench}}}
& POPE~\cite{li2023evaluating} & Hallucination probing & Contrastive image--question pairs to test visual grounding and robustness against object hallucination \\
& MMBench (MMB/MMB$^{CN}$)~\cite{liu2024mmbench} & Multi-choice & Classification-style questions covering perception, reasoning, and instruction following (English/Chinese subsets) \\
& SEED-Bench (SEED)~\cite{li2023seed} & Multimodal understanding & Vision-centric instruction-following prompts requiring image understanding, captioning, and grounding \\
& SEED$^{I}$~\cite{li2023seed} & Multi-choice & Multiple-choice questions on images, testing vision-centric understanding and instruction compliance \\
& LLaVA-Wild (LLaVA$^{W}$)~\cite{liu2023visual} & Open-ended QA & Open-ended multimodal dialogue, complex instruction following, and response quality \\
& MM-Vet~\cite{yu2023mm} & Integrated capability & Normalized score over six core vision-language skills, probing alignment and reasoning consistency \\
\bottomrule
\end{tabularx}
\endgroup
\end{table*}


The diversity of these benchmarks ensures comprehensive coverage from local details to global semantics, and from perception to reasoning, thereby thoroughly validating the adaptive capability of \name's dynamic gated pathways under varying visual granularity requirements.

\section{Hyperparameter Sensitivity Analysis}
\label{sec:appendix_hyperparam}


This section analyzes the design choices of key hyperparameters in \name, including the entropy regularization coefficient $\lambda$, the routing MLP hidden dimension $h$, and the gating initialization value $\gamma_{\mathrm{init}}$.

\textbf{Entropy Regularization Coefficient $\lambda$.}
The entropy regularizer discourages routing collapse onto a small subset of visual layers. We adopt the same stage-wise schedule as TGIF~\cite{lin2026text}: $\lambda=0.01$ during pre-training to encourage exploration across layers, and no entropy regularization during supervised fine-tuning (equivalently $\lambda=0$) so routing can specialize to downstream tasks.

\textbf{Routing MLP Hidden Dimension $h$.}
The routing MLP maps the 4096-dimensional routing feature to a 24-dimensional layer weight distribution. To balance expressiveness and computational efficiency, we set the intermediate hidden dimension to $h=1024$, aligning with the ViT hidden dimension. This design ensures that the routing network has sufficient nonlinear modeling capacity while avoiding the additional computational overhead of using the full LLM dimension (4096). 

\textbf{Gating Initialization $\gamma_{\mathrm{init}}$.}
The gating coefficient $\gamma_l=\sigma(a_l)$ controls the injection strength of visual information at the $l$-th layer, where $a_l$ is a learnable scalar. We use a default gate initialization of $\gamma_{\mathrm{init}}=0.3$ and set the gate learning rate to $5\times10^{-3}$. An initialization value that is too small leads to insufficient visual injection, causing the model to degenerate toward pure language model behavior; an initialization value that is too large may introduce excessive noise in early training, affecting convergence stability. Table~\ref{tab:gate_init} shows the impact of different initialization values on performance.

\begin{table*}[!htbp]
\centering
\caption{\textbf{Effect of gating initialization $\gamma_{\mathrm{init}}$.} We vary the initial value of the residual gate and report performance across the full benchmark suite.}
\label{tab:gate_init}
\vspace{0.2cm}
\setlength{\tabcolsep}{2.0pt}
\renewcommand{\arraystretch}{1.08}
\small
\resizebox{0.96\textwidth}{!}{
\begin{tabular}{@{}c|cccc|ccccccc@{}}
\toprule
\multirow{2}{*}{\textbf{\ensuremath{\displaystyle\gamma_{\mathrm{init}}}}}
& \multicolumn{4}{c|}{\textbf{\textit{Academic-oriented VQA}}}
& \multicolumn{7}{c}{\textbf{\textit{LMM benchmarks}}} \\
& \textbf{GQA} & \textbf{SQA$^{I}$} & \textbf{VQA$^{v2}$} & \textbf{VizWiz}
& \textbf{SEED} & \textbf{SEED$^{I}$} & \textbf{LLaVA$^{W}$} & \textbf{MM-Vet} & \textbf{MMB} & \textbf{MMB$^{CN}$} & \textbf{POPE} \\
\midrule
0.1 & 64.4 & 70.6 & 79.3 & 52.7 & 61.4 & 67.3 & 66.2 & 33.5 & 66.2 & 60.2 & \textbf{87.2} \\
\rowcolor[RGB]{240,248,255}
0.3 (default) & \textbf{64.6} & \textbf{71.0} & \textbf{79.7} & \textbf{53.1} & \textbf{61.5} & \textbf{67.5} & \textbf{66.5} & 33.8 & \textbf{66.7} & \textbf{60.5} & 87.0 \\
0.5 & 64.3 & 70.8 & 79.1 & 53.0 & 61.4 & 67.1 & 66.3 & \textbf{34.0} & 66.5 & 60.4 & 86.9 \\
\bottomrule
\end{tabular}
}
\end{table*}


The results indicate that $\gamma_{\mathrm{init}}=0.3$ achieves the best performance on most benchmarks, while $\gamma_{\mathrm{init}}=0.1$ peaks slightly on POPE and $\gamma_{\mathrm{init}}=0.5$ peaks slightly on MM-Vet. Too-small gates risk insufficient visual injection, whereas overly large gates may inject noisy visual signals early and mildly hurt broader reasoning benchmarks.

\section{Efficiency Analysis}
\label{sec:appendix_efficiency}

A natural concern is whether the dynamic gated pathways introduce prohibitive overhead. Table~\ref{tab:dgp_breakdown} breaks down the parameters of the \name module for the 7B and 13B models.

\begin{table}[!htbp]
\centering
\caption{Parameter breakdown of the \name module.}
\label{tab:dgp_breakdown}
\vspace{0.2cm}
\setlength{\tabcolsep}{4pt}
\renewcommand{\arraystretch}{1.08}
\small
\begin{tabular}{@{}l|cc|cc@{}}
\toprule
\textbf{Component} & \textbf{7B (M)} & \textbf{\% of DGP} & \textbf{13B (M)} & \textbf{\% of DGP} \\
\midrule
Routers (per-layer MLP) & 135.3 & 37.80\% & 211.2 & 37.88\% \\
Shared Cross-Attention & 67.1 & 18.75\% & 104.9 & 18.81\% \\
CLS Expert & 4.2 & 1.17\% & 5.3 & 0.95\% \\
Post-Fusion Experts & 134.4 & 37.55\% & 210.0 & 37.66\% \\
Others (embeddings, gates, norms) & 16.9 & 4.72\% & 26.2 & 4.70\% \\
\midrule
\textbf{Total DGP Module} & \textbf{357.9} & \textbf{100\%} & \textbf{557.6} & \textbf{100\%} \\
\bottomrule
\end{tabular}
\end{table}

The routers and post-fusion experts constitute the majority of additional parameters ($\sim$75\%), while the shared cross-attention module accounts for only $\sim$19\%, demonstrating efficient parameter sharing across layers.

\section{Additional Analysis}
\label{sec:appendix_qualitative}
\label{sec:appendix_extended_exp}

\subsection{Effect of Unfreezing Vision Encoder}
A natural question is whether unfreezing the vision encoder during training could further improve performance. Table~\ref{tab:unfreeze_vit} compares \name with frozen versus unfrozen ViT. Interestingly, unfreezing the ViT leads to performance drops on most benchmarks (e.g., GQA, VQA$^{v2}$,  VizWiz, SEED$^{I}$ , POPE), while only MMB shows improvement.

We attribute this to a key property of our method: the dynamic routing mechanism relies on the \emph{stable semantic hierarchy} across ViT layers---shallow layers encode textures and edges, while deeper layers capture high-level semantics. This layerwise division of labor is established during CLIP's large-scale pre-training on 400M image-text pairs. When the ViT is fine-tuned on our comparatively smaller dataset, this hierarchical structure may drift, causing the router to ``chase a moving target.'' In contrast, freezing the ViT provides a stable visual foundation, allowing the router and fusion modules to learn consistent layer-selection strategies. 

\begin{table*}[!htbp]
\centering
\caption{\textbf{Effect of unfreezing the vision encoder.} Comparison between \name with frozen and unfrozen ($\ddagger$) vision encoder during training.}
\label{tab:unfreeze_vit}
\vspace{0.2cm}
\setlength{\tabcolsep}{1.4pt}
\renewcommand{\arraystretch}{1.1}
\small
\resizebox{0.98\textwidth}{!}{%
\begin{tabular}{@{}cc|cccc|ccccccc@{}}
\toprule
\multicolumn{2}{c|}{}
& \multicolumn{4}{c|}{\textit{Academic-oriented VQA}}
& \multicolumn{7}{c}{\textit{LMM benchmarks}} \\
Method & LLM
& GQA & SQA$^{I}$ & VQA$^{v2}$ & VizWiz
& SEED & SEED$^{I}$ & LLaVA$^{W}$ & MM-Vet & MMB & MMB$^{CN}$ & POPE \\
\midrule
LLaVA-1.5~\cite{liu2024improved} & Vicuna-7B 
& 62.0 
& 66.8 
& 78.5 
& 50.0 
& 58.6 
& \textcolor{blue}{66.1} 
& 63.4 
& 30.5 
& 64.3 
& 58.3 
& \textcolor{blue}{85.9} \\

\name & Vicuna-7B 
& \textcolor{red}{64.6} 
& \textcolor{red}{71.0} 
& \textcolor{red}{79.7} 
& \textcolor{red}{53.1} 
& \textcolor{red}{61.5} 
& \textcolor{red}{67.5} 
& \textcolor{red}{66.5} 
& \textcolor{red}{33.8} 
& \textcolor{blue}{66.7} 
& \textcolor{red}{60.5} 
& \textcolor{red}{87.0} \\

\name$^{\ddagger}$ & Vicuna-7B 
& \textcolor{blue}{64.1} 
& \textcolor{red}{71.0} 
& \textcolor{blue}{78.9} 
& \textcolor{blue}{50.2} 
& \textcolor{blue}{59.7} 
& 65.3 
& \textcolor{blue}{63.5} 
& \textcolor{blue}{33.7} 
& \textcolor{red}{68.3} 
& \textcolor{blue}{59.4} 
& 84.8 \\
\bottomrule
\end{tabular}%
}
\end{table*}

\subsection{Full Routing Weight Matrix}

Table~\ref{tab:routing_matrix_pope} presents the complete routing weight matrix on the POPE benchmark, showing the average attention each LLM layer assigns to each ViT layer. Values are percentages (summing to 100 per row). Several patterns emerge: (1) different LLM layers exhibit distinct preferences---e.g., LLM layer~3 concentrates 58.8\% on ViT layer~13, while LLM layer~19 spreads mass across shallow layers 1--3; (2) middle ViT layers (7--13) receive high weights from many LLM layers, suggesting they provide a useful balance between low-level details and high-level semantics; (3) the routing distribution is highly non-uniform, confirming that \name learns meaningful layer-selection strategies rather than degenerating into uniform averaging.

\begin{table*}[t]
\centering
\caption{\textbf{Full routing weight matrix on POPE.} Each cell shows the percentage weight ($\times 100$) that LLM layer $l$ assigns to ViT layer $v$. Rows sum to 100.}
\label{tab:routing_matrix_pope}
\vspace{0.2cm}
\setlength{\tabcolsep}{1.2pt}
\renewcommand{\arraystretch}{0.92}
\tiny
\resizebox{\textwidth}{!}{%
\begin{tabular}{@{}c|cccccccccccccccccccccccc@{}}
\toprule
\textbf{LLM} & \multicolumn{24}{c}{\textbf{ViT Layer}} \\
\textbf{Layer} & \textbf{1} & \textbf{2} & \textbf{3} & \textbf{4} & \textbf{5} & \textbf{6} & \textbf{7} & \textbf{8} & \textbf{9} & \textbf{10} & \textbf{11} & \textbf{12} & \textbf{13} & \textbf{14} & \textbf{15} & \textbf{16} & \textbf{17} & \textbf{18} & \textbf{19} & \textbf{20} & \textbf{21} & \textbf{22} & \textbf{23} & \textbf{24} \\
\midrule
1  & 2.4 & 2.1 & 1.9 & 2.9 & 16.8 & 2.2 & \cellcolor{blue!15}38.3 & 2.7 & 5.8 & 6.8 & 2.8 & 2.2 & 1.8 & 1.5 & 1.6 & 1.4 & 1.3 & 1.1 & 0.9 & 0.8 & 0.7 & 0.6 & 0.6 & 0.6 \\
2  & 2.0 & 2.0 & 1.8 & 2.6 & 2.3 & 2.3 & 2.7 & 4.6 & 4.8 & 5.1 & 6.3 & 4.0 & \cellcolor{blue!12}19.1 & 12.1 & 10.2 & 9.5 & 2.8 & 1.6 & 1.0 & 0.8 & 0.7 & 0.6 & 0.5 & 0.6 \\
3  & 1.7 & 1.5 & 1.3 & 1.3 & 1.4 & 1.3 & 1.5 & 1.5 & 1.8 & 1.9 & 2.1 & 11.0 & \cellcolor{blue!22}58.8 & 2.5 & 2.5 & 1.7 & 1.4 & 1.1 & 1.0 & 0.7 & 0.6 & 0.5 & 0.4 & 0.5 \\
4  & 1.8 & 1.8 & 1.7 & 1.9 & 1.9 & 2.0 & 3.0 & 3.5 & 4.3 & 2.9 & 3.4 & 11.6 & \cellcolor{blue!18}44.9 & 2.6 & 3.3 & 2.7 & 1.7 & 1.2 & 1.0 & 0.8 & 0.6 & 0.5 & 0.4 & 0.5 \\
5  & 1.7 & 1.7 & 1.6 & 1.6 & 1.5 & 1.6 & 1.8 & 2.0 & 3.2 & 2.6 & 4.1 & 20.8 & 11.3 & \cellcolor{blue!15}29.9 & 3.8 & 3.2 & 1.9 & 1.5 & 1.0 & 0.8 & 0.7 & 0.6 & 0.5 & 0.7 \\
6  & 1.8 & 1.8 & 1.6 & 1.7 & 1.9 & 2.0 & 4.7 & 3.7 & \cellcolor{blue!18}39.9 & 21.8 & 3.0 & 2.5 & 2.3 & 1.6 & 1.7 & 1.7 & 1.2 & 1.2 & 0.9 & 0.7 & 0.6 & 0.6 & 0.6 & 0.6 \\
7  & 1.4 & 1.3 & 1.1 & 1.3 & 1.5 & 1.7 & 2.3 & 3.3 & 5.3 & 6.1 & 3.1 & 21.6 & \cellcolor{blue!15}30.8 & 4.9 & 3.2 & 2.1 & 2.5 & 1.8 & 1.5 & 0.9 & 0.7 & 0.6 & 0.5 & 0.7 \\
8  & 1.9 & 1.9 & 1.9 & 2.2 & 2.2 & 2.8 & 3.2 & 4.8 & 4.4 & 11.5 & 6.1 & \cellcolor{blue!18}34.6 & 8.7 & 2.9 & 2.3 & 1.8 & 1.6 & 1.2 & 0.9 & 0.8 & 0.7 & 0.5 & 0.5 & 0.6 \\
9  & 2.6 & 2.4 & 2.1 & 2.8 & 3.5 & 12.5 & \cellcolor{blue!18}43.8 & 3.5 & 3.1 & 6.5 & 2.4 & 2.2 & 2.0 & 1.5 & 1.5 & 1.4 & 1.2 & 1.0 & 0.9 & 0.8 & 0.6 & 0.6 & 0.5 & 0.6 \\
10 & 1.7 & 1.5 & 1.5 & 1.7 & 1.6 & 1.7 & 2.1 & 2.1 & 19.5 & \cellcolor{blue!18}44.5 & 4.7 & 3.6 & 3.0 & 1.9 & 1.5 & 1.3 & 1.1 & 1.0 & 0.8 & 0.8 & 0.7 & 0.6 & 0.6 & 0.6 \\
11 & 1.8 & 2.0 & 1.8 & 2.2 & 2.0 & 2.2 & 3.5 & 4.2 & \cellcolor{blue!15}28.3 & 5.9 & 7.5 & 11.7 & 13.7 & 2.4 & 2.2 & 1.7 & 1.3 & 1.1 & 0.9 & 0.9 & 0.8 & 0.6 & 0.6 & 0.6 \\
12 & 1.9 & 2.1 & 2.3 & 2.8 & 2.5 & 3.3 & 11.6 & 5.3 & 18.9 & 4.4 & \cellcolor{blue!15}26.1 & 3.5 & 3.7 & 1.9 & 1.7 & 1.6 & 1.2 & 1.0 & 0.9 & 0.8 & 0.7 & 0.6 & 0.6 & 0.7 \\
13 & 1.5 & 1.6 & 1.5 & 1.7 & 1.8 & 1.9 & 2.2 & 2.5 & 3.3 & 2.9 & 3.3 & 31.4 & \cellcolor{blue!18}33.9 & 1.8 & 1.4 & 1.2 & 1.1 & 0.9 & 0.8 & 0.8 & 0.7 & 0.6 & 0.6 & 0.6 \\
14 & 1.6 & 1.6 & 1.6 & 2.2 & 2.4 & 2.5 & 22.4 & 5.0 & \cellcolor{blue!15}26.0 & 8.1 & 10.4 & 2.6 & 2.3 & 1.6 & 1.4 & 1.4 & 1.3 & 1.1 & 1.0 & 0.9 & 0.8 & 0.7 & 0.7 & 0.7 \\
15 & 1.8 & 1.9 & 1.9 & 2.6 & 2.8 & 2.6 & 2.5 & 15.7 & 14.0 & \cellcolor{blue!18}36.1 & 2.8 & 2.5 & 2.3 & 1.6 & 1.2 & 1.2 & 1.0 & 0.9 & 0.8 & 0.8 & 0.8 & 0.7 & 0.7 & 0.7 \\
16 & 1.6 & 1.4 & 1.6 & 1.5 & 1.7 & 2.0 & 1.9 & 2.7 & 2.1 & 3.0 & 5.9 & 3.0 & 5.5 & 10.0 & \cellcolor{blue!18}37.9 & 7.6 & 3.8 & 2.5 & 1.3 & 0.9 & 0.7 & 0.6 & 0.4 & 0.5 \\
17 & 2.3 & 1.1 & 1.2 & 1.1 & 1.0 & 1.2 & 1.2 & 1.1 & 1.2 & 2.1 & 1.8 & 4.6 & 1.6 & \cellcolor{blue!18}38.6 & 14.9 & 8.7 & 4.8 & 4.1 & 2.5 & 2.1 & 0.8 & 0.7 & 0.6 & 0.7 \\
18 & 2.1 & 2.5 & 3.6 & 8.8 & 10.6 & 5.9 & 2.1 & 3.1 & 3.9 & 3.2 & 5.9 & \cellcolor{blue!12}17.0 & 8.1 & 2.3 & 5.3 & 5.0 & 2.7 & 3.6 & 1.3 & 0.8 & 0.7 & 0.5 & 0.3 & 0.7 \\
19 & \cellcolor{blue!15}26.2 & 12.6 & 17.2 & 5.2 & 4.0 & 2.9 & 1.7 & 3.0 & 2.5 & 1.6 & 1.7 & 3.9 & 2.8 & 3.5 & 1.7 & 1.6 & 1.4 & 1.4 & 1.0 & 0.8 & 0.5 & 0.6 & 0.5 & 1.6 \\
20 & 12.1 & \cellcolor{blue!12}17.5 & 11.9 & 7.5 & 8.7 & 4.0 & 1.9 & 2.3 & 1.4 & 2.0 & 2.9 & 8.8 & 4.1 & 2.0 & 2.9 & 1.6 & 1.9 & 1.6 & 1.4 & 1.1 & 0.5 & 0.4 & 0.9 & 0.7 \\
21 & 4.9 & 4.6 & 3.1 & 2.5 & 2.0 & 1.6 & 2.0 & 1.8 & 1.9 & 1.1 & 6.5 & 2.2 & 2.0 & 16.8 & \cellcolor{blue!18}35.1 & 3.7 & 2.0 & 1.3 & 1.2 & 0.9 & 0.9 & 0.6 & 0.5 & 0.5 \\
22 & 4.5 & 3.3 & 1.6 & 1.7 & 2.0 & 3.3 & 1.1 & 2.0 & 8.1 & 8.2 & 1.7 & 2.9 & 1.8 & 2.2 & 2.4 & 4.6 & 3.7 & \cellcolor{blue!18}36.3 & 4.6 & 1.0 & 1.5 & 0.3 & 0.4 & 0.5 \\
23 & 2.3 & 1.7 & 1.8 & 2.2 & 2.0 & 1.8 & 2.5 & 3.9 & \cellcolor{blue!15}25.3 & 9.8 & 7.0 & 4.2 & 9.4 & 10.4 & 4.3 & 4.3 & 1.2 & 1.5 & 0.9 & 0.7 & 0.5 & 0.6 & 0.8 & 0.9 \\
24 & 4.4 & 3.1 & 3.9 & \cellcolor{blue!22}53.8 & 9.0 & 2.7 & 3.4 & 1.7 & 1.6 & 1.7 & 2.6 & 1.6 & 1.0 & 1.1 & 0.7 & 0.9 & 0.7 & 0.8 & 0.7 & 0.7 & 0.7 & 0.7 & 1.1 & 1.1 \\
25 & 0.9 & 0.9 & 0.8 & 0.8 & 0.8 & 0.9 & 0.7 & 0.8 & 1.3 & 0.9 & 0.8 & 1.7 & 6.4 & \cellcolor{blue!25}72.5 & 1.7 & 1.9 & 1.4 & 0.8 & 0.7 & 0.7 & 0.6 & 0.8 & 0.6 & 0.5 \\
26 & 1.9 & 1.7 & 1.3 & 1.0 & 1.4 & 0.9 & 6.2 & 8.2 & 4.6 & 13.3 & 7.9 & \cellcolor{blue!12}13.9 & 9.6 & 8.5 & 2.2 & 2.7 & 4.1 & 3.3 & 2.6 & 1.4 & 0.7 & 0.4 & 0.6 & 1.5 \\
27 & 1.2 & 1.2 & 1.3 & 1.7 & 3.1 & 3.3 & 4.7 & 2.5 & 11.0 & \cellcolor{blue!22}54.0 & 2.9 & 1.8 & 1.8 & 1.0 & 0.8 & 1.0 & 0.8 & 0.9 & 0.7 & 0.8 & 0.6 & 0.8 & 0.7 & 1.4 \\
28 & 1.9 & 1.7 & 1.3 & 2.5 & 1.5 & 2.3 & 3.4 & \cellcolor{blue!25}62.6 & 3.4 & 2.7 & 3.5 & 2.4 & 2.4 & 1.5 & 1.0 & 0.7 & 0.5 & 0.6 & 0.7 & 0.8 & 0.6 & 0.6 & 0.5 & 0.7 \\
29 & 1.1 & 1.1 & 1.1 & 1.5 & 1.3 & 1.1 & 1.1 & 1.0 & 1.6 & \cellcolor{blue!22}54.6 & 12.2 & 5.7 & 3.7 & 3.3 & 1.2 & 1.1 & 1.1 & 1.0 & 1.2 & 1.2 & 0.8 & 0.7 & 0.7 & 0.6 \\
30 & 1.5 & 1.4 & 0.8 & 1.1 & 1.0 & 0.5 & 0.8 & 0.9 & 0.9 & 1.4 & 1.3 & 1.0 & 1.1 & 1.3 & 11.8 & 4.9 & \cellcolor{blue!22}50.0 & 7.0 & 4.0 & 1.3 & 1.7 & 1.6 & 1.8 & 0.9 \\
31 & 2.9 & 2.3 & 1.5 & 1.2 & 1.0 & 1.1 & 1.4 & 2.1 & 6.9 & \cellcolor{blue!22}51.1 & 10.3 & 3.5 & 3.6 & 2.2 & 1.2 & 1.1 & 0.8 & 0.6 & 1.1 & 0.9 & 0.8 & 0.8 & 0.7 & 1.0 \\
32 & 3.4 & 1.0 & 3.4 & 7.8 & 6.4 & 1.6 & 2.7 & 1.0 & 2.6 & 3.8 & 1.5 & 0.4 & 0.4 & 9.6 & 14.1 & \cellcolor{blue!12}20.2 & 2.2 & 0.8 & 1.6 & 1.0 & 1.7 & 4.2 & 6.0 & 2.6 \\
\bottomrule
\end{tabular}%
}
\end{table*}

\subsection{Quantitative Evidence for Routing Patterns}

To provide causal evidence that the learned routing patterns are functionally meaningful, we conduct inference-time intervention experiments on the trained DGP-7B model \emph{without retraining}.

\textbf{Experimental Setup.} We compare four routing strategies: (1) \textbf{DGP (Learned)}: the learned 24-dim routing weights; (2) \textbf{Baseline (Last-2)}: LLaVA-1.5 default using only the penultimate ViT layer; (3) \textbf{Uniform}: equal weights across all 24 ViT layers; (4) \textbf{Reversed}: flipping the learned routing weights while retaining access to all ViT layers.

\begin{table*}[!htbp]
\centering
\caption{\textbf{Inference-time routing intervention.} Comparison of different routing strategies on DGP-7B without retraining.}
\label{tab:routing_intervention}
\vspace{0.2cm}
\setlength{\tabcolsep}{1.5pt}
\renewcommand{\arraystretch}{1.08}
\small
\resizebox{\textwidth}{!}{%
\begin{tabular}{@{}l|cccc|ccccccc@{}}
\toprule
\multirow{2}{*}{\textbf{Routing}} & \multicolumn{4}{c|}{\textbf{\textit{Academic-oriented VQA}}} & \multicolumn{7}{c}{\textbf{\textit{LMM benchmarks}}} \\
& \textbf{GQA} & \textbf{SQA$^{I}$} & \textbf{VQA$^{v2}$} & \textbf{VizWiz} & \textbf{SEED} & \textbf{SEED$^{I}$} & \textbf{LLaVA$^{W}$} & \textbf{MM-Vet} & \textbf{MMB} & \textbf{MMB$^{CN}$} & \textbf{POPE} \\
\midrule
DGP (Learned) & \textbf{64.6} & \textbf{71.0} & \textbf{79.7} & \textbf{53.1} & \textbf{61.5} & \textbf{67.5} & \textbf{66.5} & \textbf{33.8} & \textbf{66.7} & \textbf{60.5} & \textbf{87.0} \\
Baseline (Last-2) & 62.0 & 66.8 & 78.5 & 50.0 & 58.6 & 66.1 & 63.4 & 30.5 & 64.3 & 58.3 & 85.9 \\
Uniform & 60.2 & 64.3 & 72.0 & 50.2 & 57.9 & 63.3 & 56.3 & 25.3 & 57.6 & 44.9 & 77.2 \\
Reversed & 63.6 & 67.5 & 77.8 & 52.4 & 61.3 & 67.1 & 57.0 & 32.1 & 61.7 & 55.8 & 86.6 \\
\bottomrule
\end{tabular}%
}
\end{table*}

\textbf{Findings.} As shown in Table~\ref{tab:routing_intervention}, (i) \emph{Learned $>$ Baseline $>$ Uniform}: Dynamic routing consistently outperforms both single-layer baseline and naive averaging, confirming the necessity of selective multi-layer fusion.
(ii) \emph{Reversed degrades but does not collapse}: Since each LLM layer still accesses the same ViT layers via its trained router, performance remains above Uniform. However, perturbing the weight distribution leads to consistent degradation (e.g., LLaVA$^{W}$: 66.5$\to$57.0, MMB: 66.7$\to$61.7).
These results demonstrate that performance depends not only on \emph{which} ViT layers are accessible, but also on \emph{how} they are weighted, providing causal evidence that the routing patterns are functionally meaningful.

\subsection{Additional Visualizations}
\begin{figure}[H]
  \centering
  \includegraphics[width=1\linewidth]{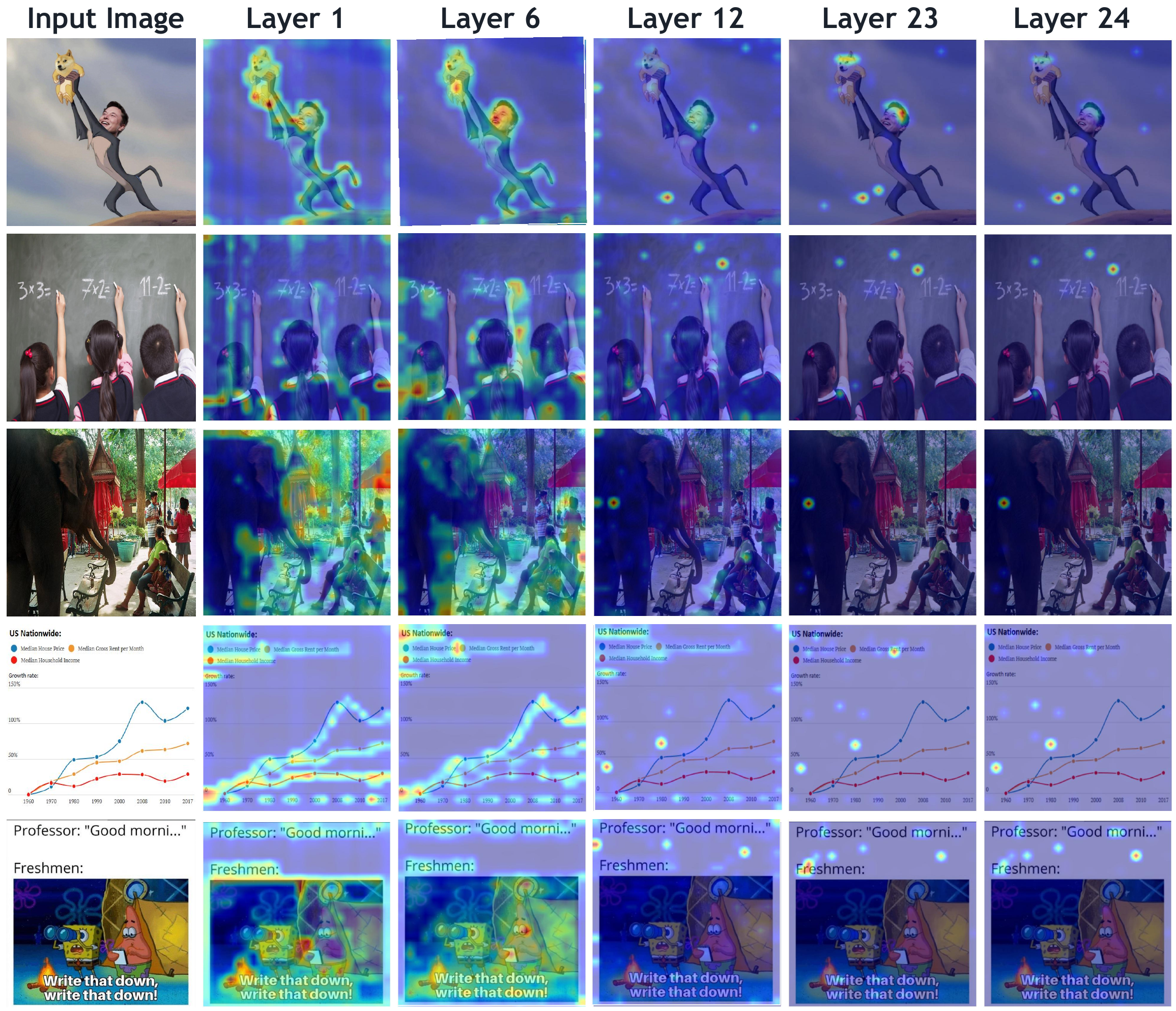}
  \vspace{-0.15in}
  \caption{\textbf{Additional attention visualizations across ViT layers.} Supplementary heatmaps showing the progression of attention patterns from shallow to deep ViT layers.}
  \label{fig:attention_supp}
\end{figure}

\begin{figure}[H]
  \centering
  \includegraphics[width=1\linewidth]{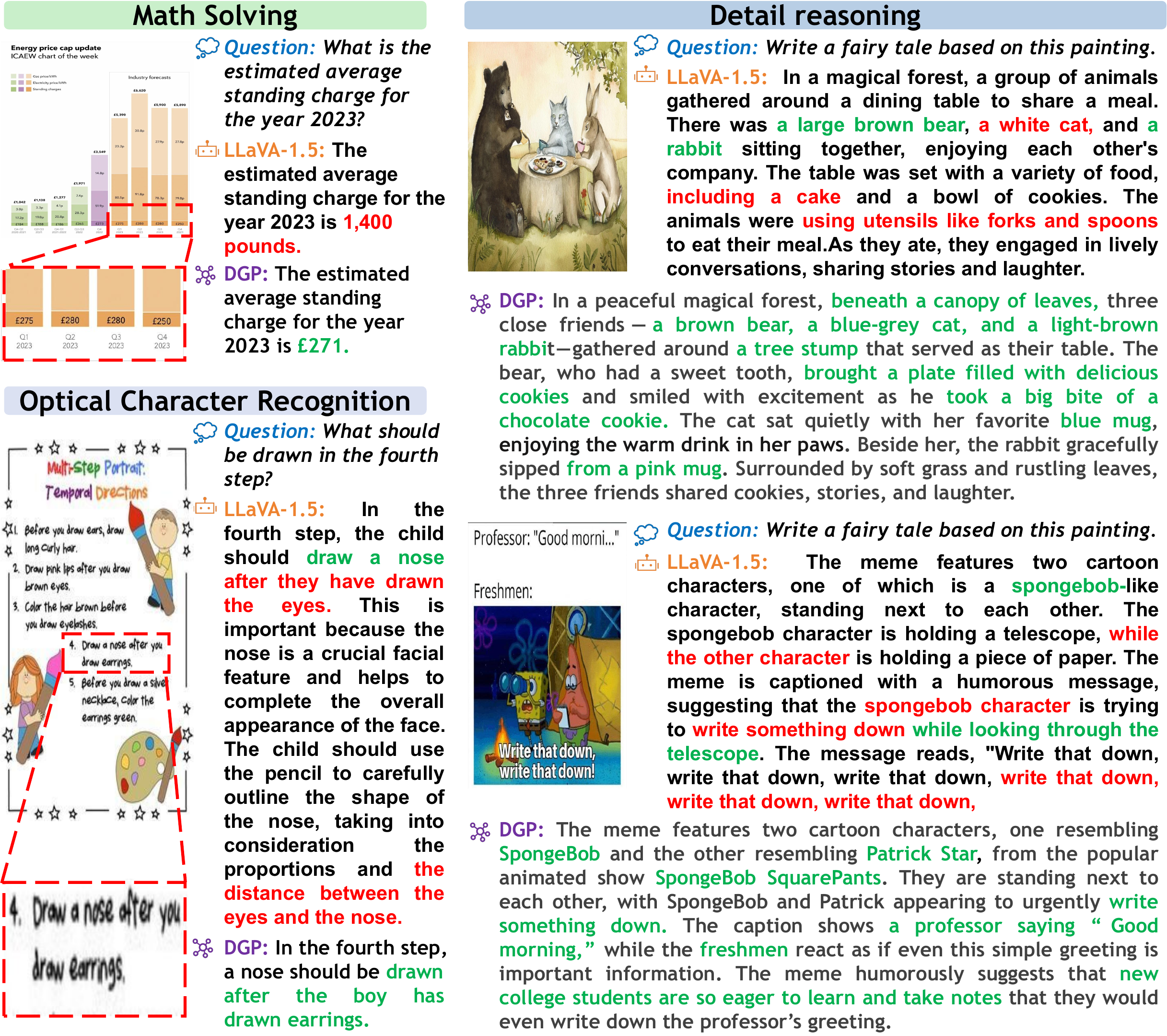}
  \vspace{-0.15in}
  \caption{\textbf{Additional qualitative comparison examples.} More case studies comparing \name with LLaVA-1.5 across diverse visual understanding tasks.}
  \label{fig:case_study_supp}
\end{figure}

\begin{figure}[H]
  \centering
  \includegraphics[width=1\linewidth]{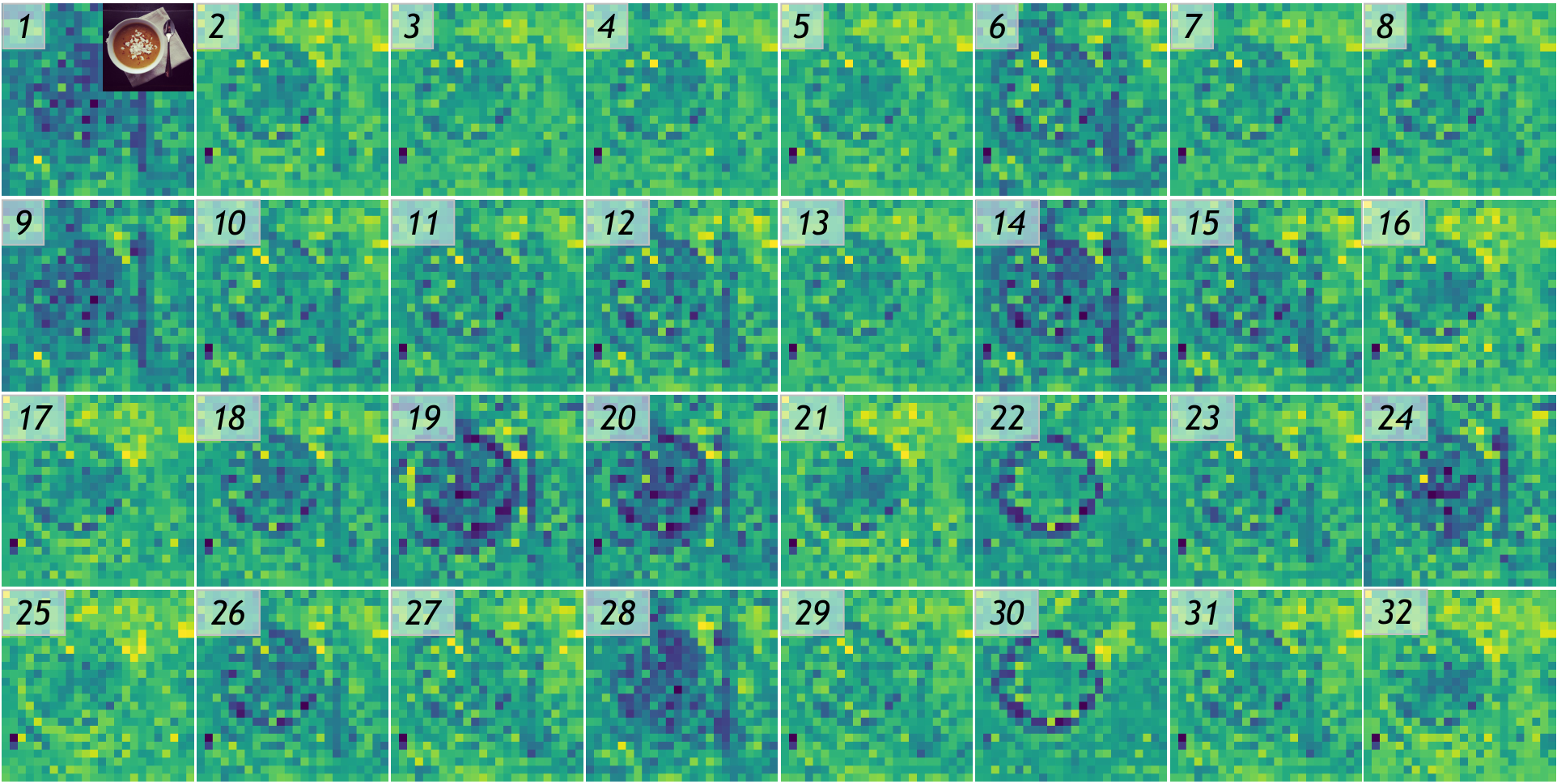}
  \vspace{-0.15in}
  \caption{\textbf{Additional feature visualizations (Part 1).} Supplementary \name gated-fusion feature maps across all 32 LLM layers, illustrating improved spatial structure and semantic coherence.}
  \label{fig:feature_supp2}
\end{figure}

\begin{figure}[H]
  \centering
  \includegraphics[width=1\linewidth]{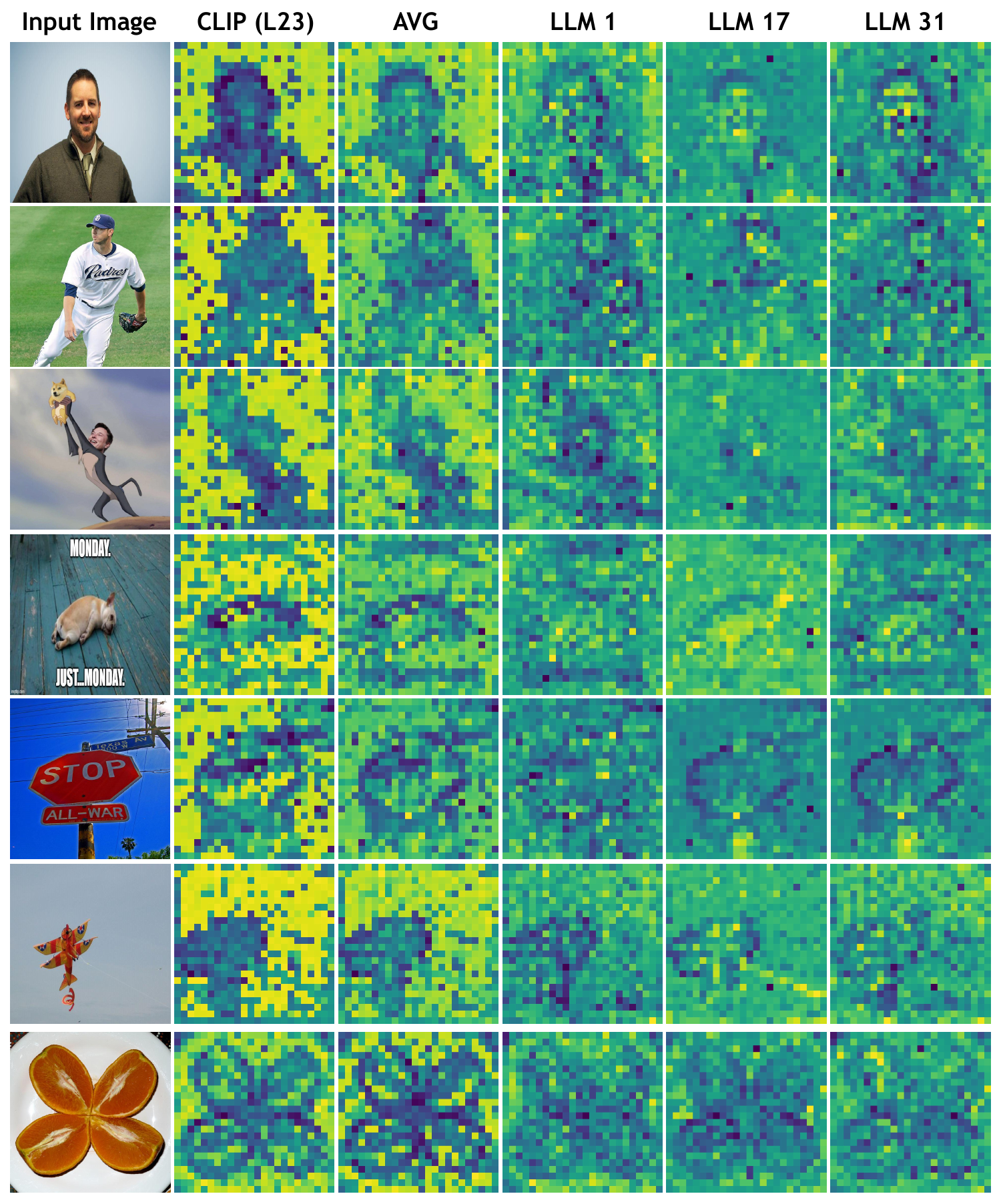}
  \vspace{-0.15in}
  \caption{\textbf{Additional feature visualizations (Part 2).} More comparisons of fixed single-layer features, uniform average fusion, and \name's gated fusion at different LLM layers.}
  \label{fig:feature_supp1}
\end{figure}

\begin{figure}[H]
  \centering
  \includegraphics[width=1\linewidth]{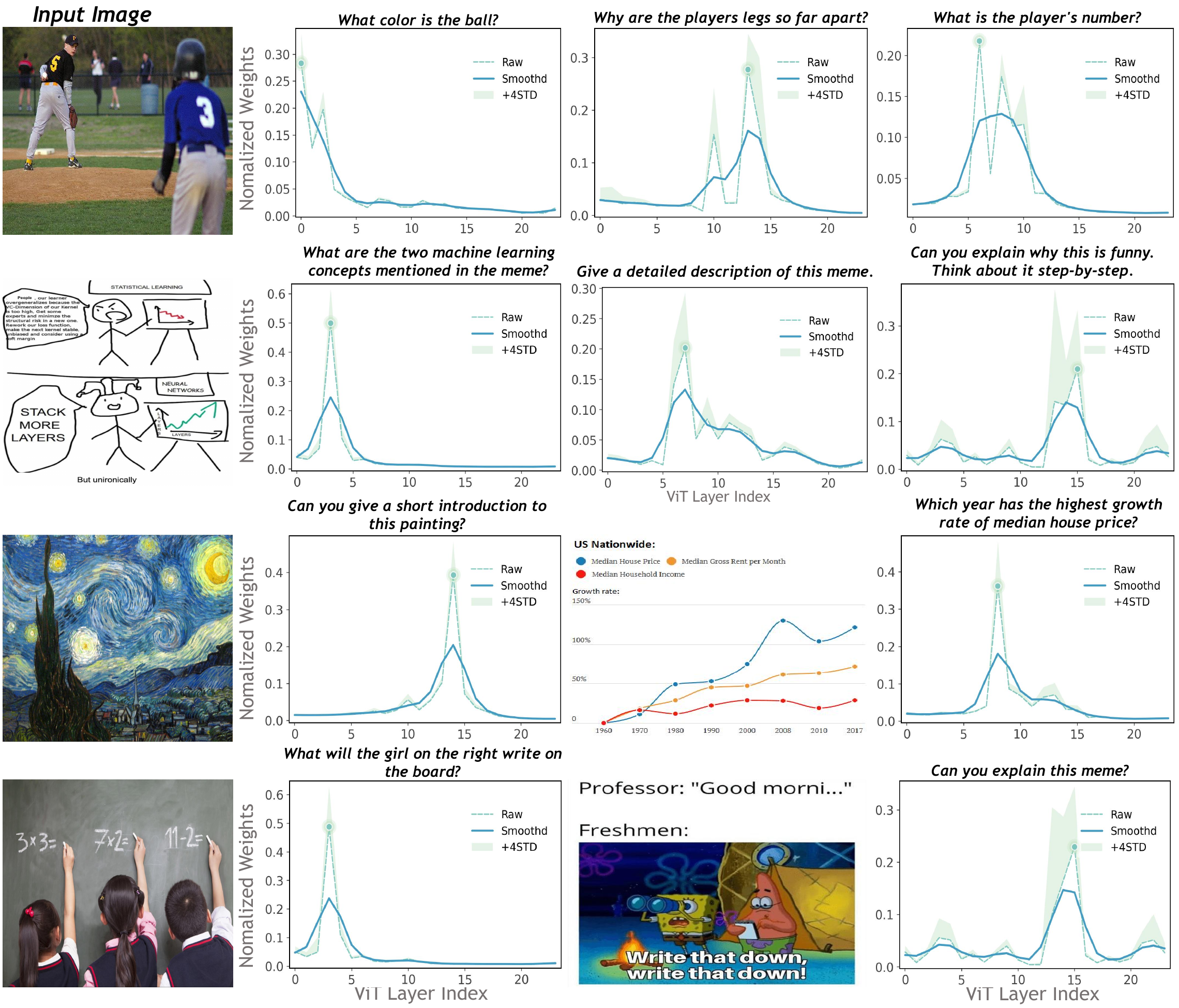}
  \vspace{-0.15in}
  \caption{\textbf{Additional question-aware routing analysis.} More examples showing how routing weights shift as the question varies for the same image.}
  \label{fig:query_supp}
\end{figure}

\begin{figure}[H]
  \centering
  \includegraphics[width=1\linewidth]{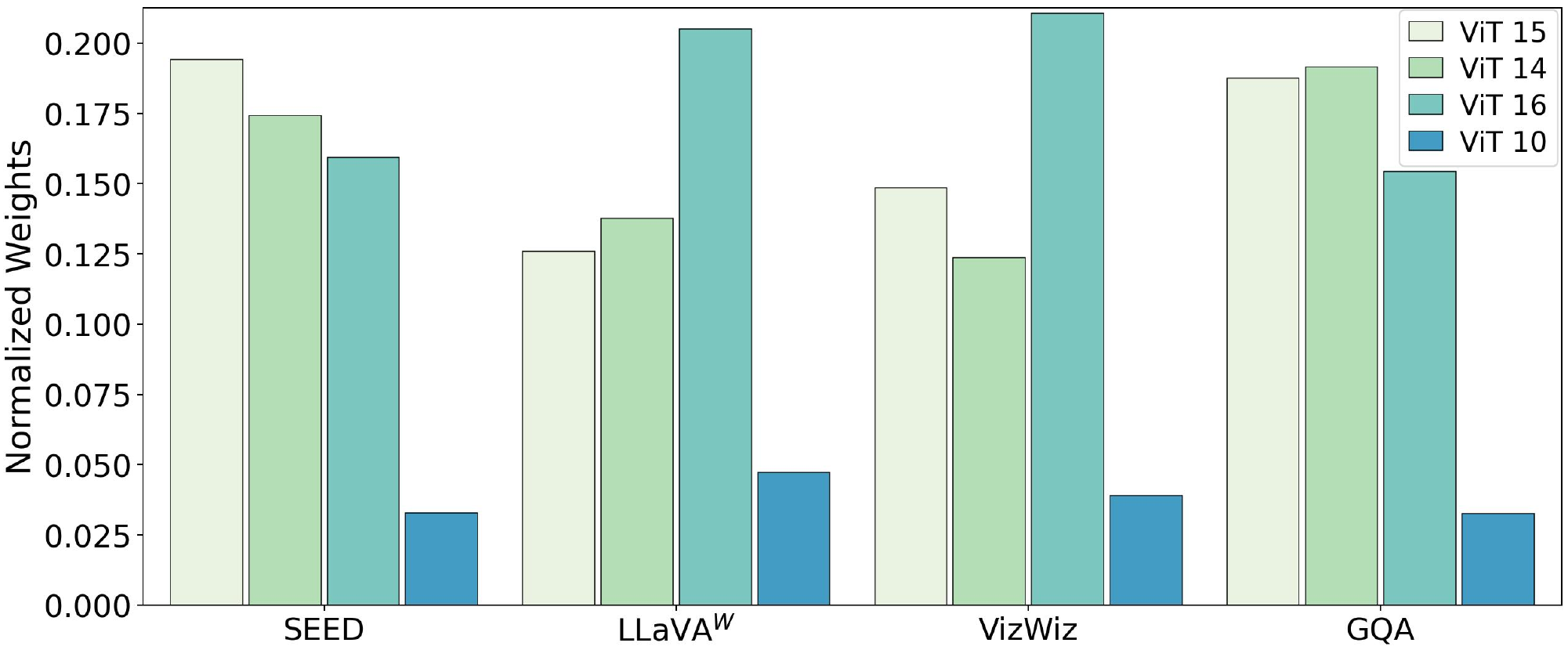}
  \vspace{-0.15in}
  \caption{\textbf{Routing weight distributions at LLM layer~32.} Weights that the last LLM layer assigns across ViT layers on different benchmarks, showing that distinct tasks favor different ViT-layer combinations.}
  \label{fig:llm32_vis}
\end{figure}

\begin{figure}[H]
  \centering
  \includegraphics[width=1\linewidth]{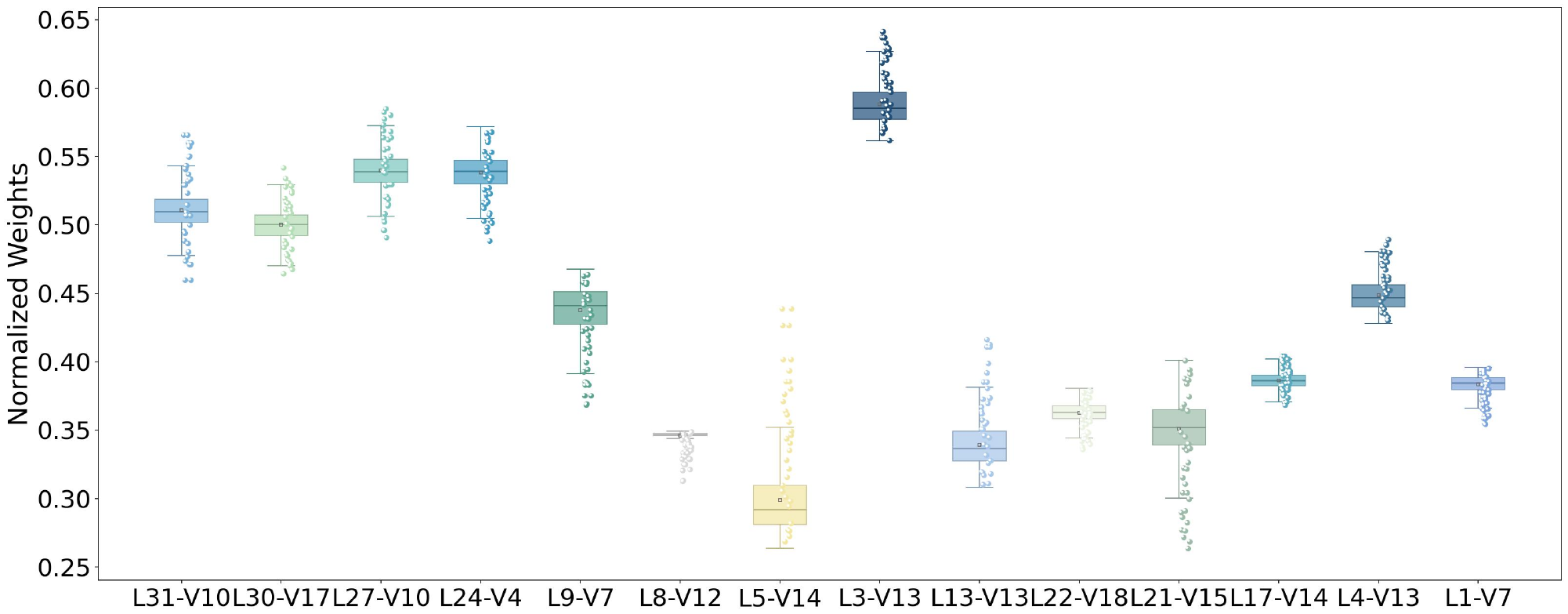}
  \vspace{-0.15in}
  \caption{\textbf{Additional routing analysis (Part 1).} Supplementary visualizations of sample-wise routing weights.}
  \label{fig:routing_supp1}
\end{figure}

\begin{figure}[H]
  \centering
  \includegraphics[width=1\linewidth]{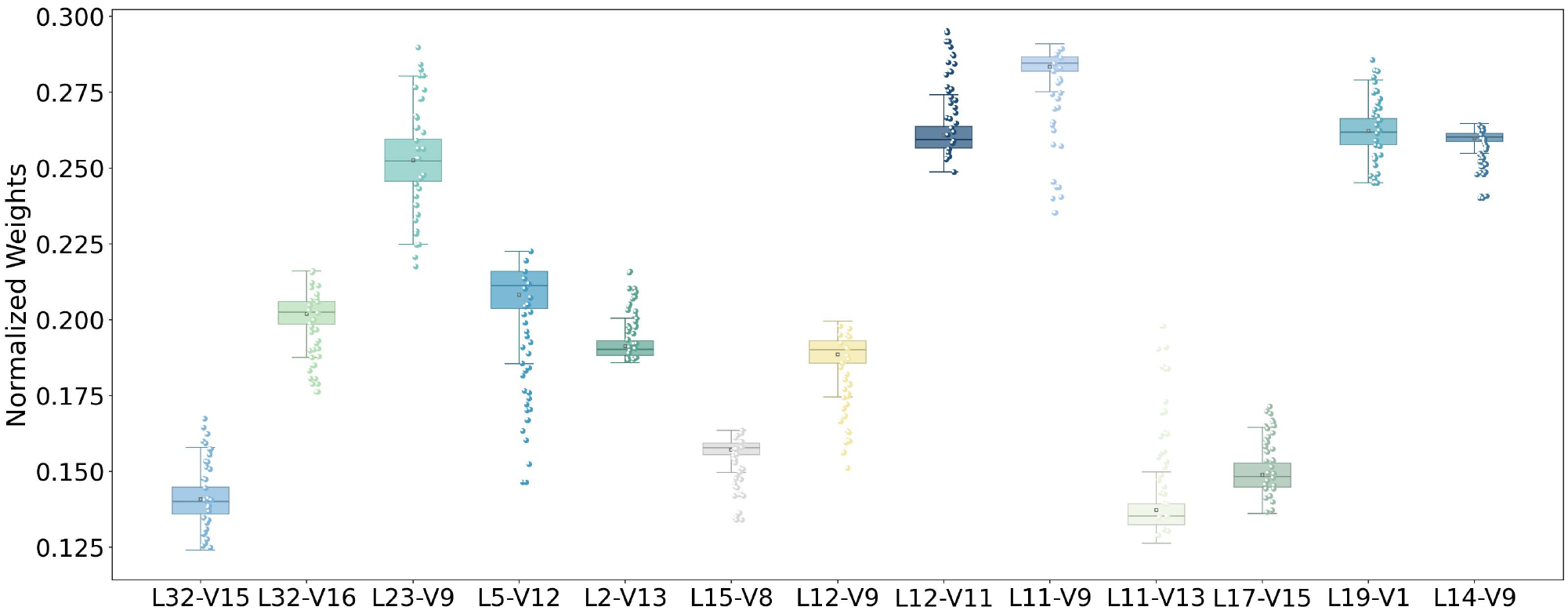}
  \vspace{-0.15in}
  \caption{\textbf{Additional routing analysis (Part 2).} More examples showing the diversity of routing patterns across different inputs.}
  \label{fig:routing_supp2}
\end{figure}

Figure~\ref{fig:attention_supp} shows additional ViT attention heatmaps illustrating how attention shifts from low-level edges to high-level semantics across layers.
Figure~\ref{fig:case_study_supp} presents more case-study comparisons between \name and LLaVA-1.5, demonstrating improved fine-grained reasoning.
Figure~\ref{fig:feature_supp2} shows \name gated-fusion feature maps across all 32 LLM layers, illustrating improved spatial structure and semantic coherence.
Figure~\ref{fig:feature_supp1} compares fixed single-layer, uniform-average, and \name gated-fusion features, showing that our method yields sharper boundaries and more coherent object segmentation.
Figure~\ref{fig:query_supp} further illustrates question-aware routing: for the same image, different questions trigger distinct routing patterns across ViT layers.
Figure~\ref{fig:llm32_vis} shows routing weight distributions at LLM layer~32 across different benchmarks, revealing that distinct tasks favor different ViT-layer combinations.
Figures~\ref{fig:routing_supp1} and~\ref{fig:routing_supp2} provide additional routing weight distributions for various samples, showing that different inputs produce different weighting patterns.

\section{Limitations and Future Work}
\label{sec:appendix_limitations}

The Dynamic Gated Pathways module is designed to be \emph{plug-and-play}: it introduces only lightweight, layer-wise routing components that interface with the LLM hidden states, without modifying the vision encoder or the LLM architecture itself. This modularity suggests that \name should generalize across vision encoders and LLM backbones, adaptively selecting appropriate visual granularities regardless of the specific encoder--decoder pairing.

In this work, we validate \name primarily under the LLaVA-1.5 setting (CLIP + Vicuna-7B/13B) and further on smaller-scale alternative frameworks, including Qwen3-VL-2B (SigLIP2 + Qwen3-1.7B) and LLaVA-NeXT (AnyRes tiling). Due to resource and time constraints, however, experiments on larger LLMs (30B+), broader families of vision encoders, and larger-scale multimodal corpora remain limited. Extending validation along these axes constitutes our primary future work. We anticipate that the dynamic routing mechanism will remain effective---and potentially more beneficial---as model capacity and data scale increase, since richer language representations should enable even finer-grained visual-layer selection.

\section{Broader Impacts}
\label{sec:appendix_broader_impacts}

This work improves hierarchical visual access in MLLMs, which may benefit applications such as accessibility tools and educational assistants. At the same time, stronger visual reasoning could be misused for surveillance or misleading content, and \name may inherit biases from the pretrained vision and language backbones it fine-tunes. We therefore view \name as an architectural improvement for research use under the licenses and safety practices of the host models.

\section{Use of Large Language Models}
\label{app:llm_writing}
For transparency, we disclose that general-purpose large language models (LLMs) were used during the preparation of this manuscript \emph{solely as writing-assistance tools} to improve the clarity, concision, and grammaticality of the text. Their use was restricted to language-level editing: rewriting sentences for readability, tightening phrasing, polishing transitions, and correcting typos. No LLM was used to generate scientific claims, design experiments, produce technical content, or decide on methodology; all results, figures, analyses, and conclusions were conceived and verified by the authors, who take full responsibility for the content of every paragraph.

\vspace{-0.25in}

\end{CJK*}

\end{document}